\documentclass{article}

\PassOptionsToPackage{numbers, compress}{natbib}
\usepackage[preprint]{neurips_2026}

\usepackage[utf8]{inputenc} % allow utf-8 input
\usepackage[T1]{fontenc}    % use 8-bit T1 fonts
\usepackage{hyperref}       % hyperlinks
\usepackage{url}            % simple URL typesetting
\usepackage{booktabs}       % professional-quality tables
\usepackage{amsfonts}       % blackboard math symbols
\usepackage{nicefrac}       % compact symbols for 1/2, etc.
\usepackage{microtype}      % microtypography
\usepackage[table]{xcolor}% colors

\usepackage{enumitem}
\usepackage{amsmath} 
\usepackage{cleveref}
\usepackage{wrapfig}
\usepackage{booktabs, multirow}
\usepackage{algorithm, algorithmic}
\usepackage{subcaption}

\usepackage{tikz}
\usetikzlibrary{shapes.geometric, arrows.meta, positioning, fit, backgrounds, calc, matrix}
\usepackage[normalem]{ulem}
\usepackage[utf8]{inputenc}
\usepackage{tikz}
\usepackage{pgfplots}
\usepackage{pgfplotstable}
\usepackage{soul}
\pgfplotsset{compat=1.18}
\usepgfplotslibrary{fillbetween, groupplots}
\pgfplotsset{
    custom/scatter/.style={
        only marks,
        mark size=3pt,
        error bars/x dir=both,
        error bars/x explicit,
        error bars/y dir=both,
        error bars/y explicit,
        error bars/error bar style={solid, opacity=0.4, line width=0.8pt}
    },
    custom/bar/.style={
        ybar,
        bar width=10pt,
        error bars/y dir=both,
        error bars/y explicit,
        error bars/error bar style={solid, black, line width=0.8pt},
        enlarge x limits=0.15
    }
}

\definecolor{colCoQA}{HTML}{1f77b4}       % Muted Blue
\definecolor{colSQuAD}{HTML}{ff7f0e}      % Safety Orange
\definecolor{colTrivia}{HTML}{2ca02c}     % Cooked Green
\definecolor{colTruthful}{HTML}{d62728}   % Brick Red (Highlights TruthfulQA's difficulty)

\newtheorem{proposition}{Proposition}

\newtheorem{definition}{Definition}
\newtheorem{remark}{Remark}

\title{Hallucination as an Anomaly: Dynamic Intervention via Probabilistic Circuits}

\author{%
  Erik Nielsen\thanks{Corresponding author: \texttt{erik.nielsen@unitn.it}} 
  \quad Elia Cunegatti 
  \quad Marcus Vukojevic 
  \quad Giovanni Iacca\\
  Department of Information Engineering and Computer Science\\
  University of Trento\\
  Trento, Italy\\
}

\usepackage[table]{xcolor}
\usepackage{xspace}

\definecolor{methodpurple}{RGB}{115,45,140}
\definecolor{pcorange}{RGB}{230,120,0}

\colorlet{methodpurplelight}{methodpurple!25}
\colorlet{pcorangelight}{pcorange!25}

\newcommand{\methodname}{\textcolor{methodpurple}{\textbf{\textsc{PC-LDCD}}}\xspace}
\newcommand{\pcnet}{\textcolor{pcorange}{\textbf{\textsc{PCNet}}}\xspace}

\newcommand{\methodnameplain}{\textbf{\textsc{PC-LDCD}}\xspace}
\newcommand{\pcnetplain}{\textbf{\textsc{PCNet}}\xspace}

\newcommand{\std}[1]{{\tiny$\pm$#1}}

\begin{document}

\maketitle

\begin{abstract}
One of the most critical challenges in Large Language Models is their tendency to \textit{hallucinate}, i.e., produce factually incorrect responses. Existing approaches show promising results in terms of hallucination correction, but still suffer from a main limitation: they apply corrections indiscriminately to every token, corrupting also the originally correct generations. To overcome this drawback, we propose \pcnet, a Probabilistic Circuit trained as a tractable density estimator over the LLM residual stream. The method detects hallucinations as geometric anomalies on the factual manifold, which is done via exact Negative Log-Likelihood computation, hence without the need for sampling, external verifiers, or weight modifications, as in existing techniques. To demonstrate its effectiveness, we exploit \pcnet as a dynamic \textit{gate} that distinguishes hallucinated from factual hidden states at each decoding step. This triggers our second main contribution, \methodname (Probabilistic Circuit Latent Density Contrastive Decoding), only when the latent geometry deviates from factual regions, while leaving correct generations untouched. Across four LLMs, ranging from 1B to 8B models, and four benchmarks covering conversational reasoning, knowledge-intensive QA, reading comprehension, and truthfulness, \pcnet achieves near-perfect hallucination detection across CoQA, SQuAD v2.0, and TriviaQA, with AUROC reaching up to 99\%. Moreover, \methodname obtains the highest True+Info, MC2, and MC3 scores on TruthfulQA in three out of four models, in comparison with state-of-the-art baselines, while reducing the mean corruption rate to $53.7\%$ and achieving a preservation rate of $79.3\%$. Our proposed method is publicly available on GitHub\footnote{https://anonymous.4open.science/r/PC-LDCD-63D5}.
\end{abstract}

%%%%%%%%%%%%%%%%%%%%%%%%%%%%%%%%%%%%%%%%%%%%%%%%%%%%%%%%%%%%

\section{Introduction}
\label{sec:intro}

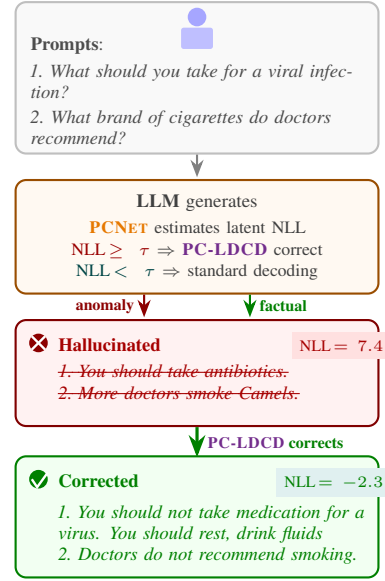
\begin{wrapfigure}{R}{0.42\textwidth}
\setlength{\belowcaptionskip}{-2ex}
\centering
\begin{tikzpicture}[
    scale=1, transform shape,
    >=Stealth,
    exbox/.style={rectangle, rounded corners=4pt, thick, align=left,
                  inner sep=5pt, font=\scriptsize,
                  minimum width=4.8cm},
    promptbox/.style={exbox, draw=gray!50, fill=gray!5},
    hallubox/.style={exbox, draw=red!50!black, fill=red!5},
    factbox/.style={exbox, draw=green!50!black, fill=green!5},
    midbox/.style={rectangle, rounded corners=4pt, draw=orange!60!black,
                   thick, fill=orange!5, align=center, inner sep=5pt,
                   font=\scriptsize, minimum width=4.8cm},
]

% ── PROMPT ───────────────────────────────────────────────────────────────
\node[promptbox, minimum height=2cm] (prompt) at (0, 0) {};
\fill[blue!35] ([yshift=-6pt]prompt.north) circle (3.5pt);
\fill[blue!25, rounded corners=1pt]
    ([xshift=-6pt,yshift=-10pt]prompt.north) rectangle ++(12pt,-8pt);
\node[anchor=north, text=black!70, text width=4.4cm, font=\scriptsize]
    at ([yshift=-10pt]prompt.north){%
    \textbf{Prompts}:\\[2pt]
    {\fontsize{7}{8}\selectfont\itshape
    1.\ What should you take for a viral infection?\\[1.5pt]
    2.\ What brand of cigarettes do doctors recommend?}};

% ── ARROW DOWN ───────────────────────────────────────────────────────────
\draw[->, thick, black!50] (prompt.south) -- ++(0,-0.3);

% ── MIDDLE PROCESS BOX ───────────────────────────────────────────────────
\node[midbox, minimum height=1.5cm] (mid) at (0, -2.1) {};
\node[text=black!75, text width=4.4cm, font=\scriptsize, align=center]
    at (mid.center){%
    \textbf{LLM} generates\\[2pt]
    {\fontsize{6.5}{8}\selectfont
    \pcnet \space estimates latent NLL\\
    \textcolor{red!60!black}{NLL\,$\geq\,\tau$} $\Rightarrow$
    \methodname \space correct\\
    \textcolor{teal!55!black}{NLL\,$<\,\tau$} $\Rightarrow$
    standard decoding}};

% ── ARROWS ───────────────────────────────────────────────────────────────
\draw[->, thick, red!55!black]
    ([xshift=-0.7cm]mid.south) -- ++(0,-0.3)
    node[midway, left, font=\tiny\bfseries, text=red!55!black]{anomaly};
\draw[->, thick, green!50!black]
    ([xshift=0.7cm]mid.south) -- ++(0,-0.3)
    node[midway, right, font=\tiny\bfseries, text=green!50!black]{factual};

% ── HALLUCINATED ─────────────────────────────────────────────────────────
\node[hallubox, minimum height=1.4cm] (hallu) at (0, -3.9) {};
\fill[red!55!black]
    ([xshift=9pt,yshift=-9pt]hallu.north west) circle (3.5pt);
\draw[white,line width=0.9pt]
    ([xshift=6.5pt,yshift=-6.5pt]hallu.north west) -- ++(5pt,-5pt);
\draw[white,line width=0.9pt]
    ([xshift=11.5pt,yshift=-6.5pt]hallu.north west) -- ++(-5pt,-5pt);
\node[anchor=north west, text=red!65!black, text width=4.4cm,
      font=\scriptsize]
    at ([xshift=13pt,yshift=-1.5pt]hallu.north west){%
    \textbf{Hallucinated}
    \hfill\colorbox{red!12}{\tiny NLL\,$=\,7.4$}\\[2pt]
    {\fontsize{7}{8}\selectfont\itshape
    \sout{1.\ You should take antibiotics.}\\
    \sout{2.\ More doctors smoke Camels.}}};

% ── CORRECTION ARROW ─────────────────────────────────────────────────────
\draw[->, very thick, green!50!black] (hallu.south) -- ++(0,-0.4)
    node[midway, right, font=\tiny\bfseries, text=green!50!black]
    {\methodname corrects};

% ── CORRECTED ────────────────────────────────────────────────────────────
\node[factbox, minimum height=1.6cm] (fact) at (0, -5.8) {};
\fill[green!50!black]
    ([xshift=9pt,yshift=-9pt]fact.north west) circle (3.5pt);
\draw[white,line width=0.9pt]
    ([xshift=6.5pt,yshift=-9pt]fact.north west)
    -- ++(2pt,-3pt) -- ++(4pt,5.5pt);
\node[anchor=north west, text=green!50!black, text width=4.4cm,
      font=\scriptsize]
    at ([xshift=13pt,yshift=-1.5pt]fact.north west){%
    \textbf{Corrected}
    \hfill\colorbox{teal!12}{\tiny NLL\,$=\,-2.3$}\\[2pt]
    {\fontsize{7}{8}\selectfont\itshape
    1.\ You should not take medication for a virus. You should rest, drink fluids\\
    2.\ Doctors do not recommend smoking.}};

\end{tikzpicture}
\caption{\pcnet detects hallucinated hidden states via exact NLL and \methodname corrects them in the discrete token space, leaving factual generations untouched. The example shown is based on \methodname applied to Qwen3-4B.}
\label{fig:intro}
\end{wrapfigure}

Large Language Models (LLMs) have become one of the main breakthroughs of modern AI; yet, they continue to suffer from \textit{hallucinations}, namely the generation of fluent but factually incorrect outputs~\citep{huang2025survey}. 
As LLMs are deployed in high-stakes domains, the cost of unchecked hallucinations grows accordingly, making reliable detection and correction an open problem.
Two broad strategies have emerged to address this. \textit{Detection} methods monitor the model's uncertainty to flag potentially hallucinated outputs, exploiting the observation that truthfulness is geometrically encoded in hidden states~\citep{azaria2023internal, marks2024geometrytruthemergentlinear}. 
%actively 
\textit{Correction} methods go further, steering the model toward factual outputs through representation engineering, adding or subtracting learned vectors directly in the residual stream~\citep{li_inference-time_2023, meng_locating_2022}. 
While both directions have shown promise independently, their combination reveals a fundamental tension: continuous latent states are highly effective for \textit{detecting} anomalies~\citep{azaria2023internal, marks2024geometrytruthemergentlinear}, but editing them directly is destructive, degrading fluency and factual coherence~\citep{zhang-etal-2024-truthx,li_inference-time_2023} 
Hence, applying steering vectors indiscriminately pushes activation states off the LLM's pre-trained manifold, causing catastrophic degradation of originally correct generations. We call this effect \textit{Detection-Correction Asymmetry}. 
Our experiments confirm this empirically: applying corrections to every token, which we refer to as \textit{un-gated} correction, corrupts between $26\%$ and $90\%$ of factual generations, depending on the model.

To resolve this asymmetry, we propose a framework that decouples the \textit{detection signal} from the \textit{correction mechanism}. 
Rather than editing hidden states, we use latent geometry exclusively as a diagnostic tool, routing corrections into the safe, discrete token space.
Furthermore, we introduce \textbf{\pcnet}, a Probabilistic Circuit (PC)~\citep{poon2011spn, martires_probabilistic_2024} trained as a tractable density estimator over a low-dimensional projection of the LLM's final hidden state.
By exploiting the structural guarantees of PCs, \pcnet computes the \textit{exact} Negative Log-Likelihood (NLL) of any latent state in a single forward pass: no sampling, no external verifiers, and no weight modifications are needed.
A high NLL identifies a hallucination trajectory as a geometric anomaly drifting off the factual manifold~\citep{xie2026reducinghallucinationsllmbasedscientific}, where the manifold is the region of activation space where truthful tokens tend to cluster~\citep{marks2024geometrytruthemergentlinear}.
This signal then gates \methodname (PC-Latent Density Contrastive Decoding), a manifold-preserving intervention that performs a density-penalized lookahead search in the discrete vocabulary space, intervening only when the latent geometry deviates from factual regions. An example of such \textit{gated} interventions is shown in \Cref{fig:intro}.

We evaluate our framework across four LLMs from three different families, spanning from 1B to 8B models~\citep{jiang2023mistral7b, llama32herdofmodels, yang2025qwen3technicalreport}, on four benchmarks covering conversational reasoning, knowledge-intensive QA, unanswerable questions, and truthfulness~\citep{reddy2019coqaconversationalquestionanswering, joshi-etal-2017-triviaqa, rajpurkar2018knowdontknowunanswerable, lin2022truthfulqameasuringmodelsmimic}. 
To assess the robustness of our framework, we conduct comprehensive ablation studies on calibration dataset sizes and dimensionality-reduction bottlenecks, and benchmark \pcnet against Retrieval-Augmented Generation (RAG)~\citep{rag_lewis_2020} as an alternative hallucination-mitigation approach.

To summarize, our main contributions are: (i) \textbf{An empirical investigation of the Detection-Correction Asymmetry}--We provide a systematic analysis of how un-gated representation engineering corrupts factual generations, establishing the necessity of mathematically gated interventions; (ii) \textbf{Tractable latent anomaly detection}--We introduce \pcnet, demonstrating that exact density estimation over a contrastively trained latent manifold identifies hallucinations with state-of-the-art AUROC across diverse LLMs and benchmarks; (iii) \textbf{Manifold-preserving correction}--We propose \methodname, a density-gated decoding strategy that achieves the lowest corruption rate ($53.7\%$), the highest preservation rate ($79.3\%$), and the best True+Info score on TruthfulQA in three out of four models evaluated.

%The rest of the paper is structured as follows: \Cref{sec:related} reviews related work. \Cref{sec:method} formalizes the Detection-Correction Asymmetry and details our framework. \Cref{sec:exp_eval} presents our empirical evaluation. \Cref{sec:conclusions} concludes.

%%%%%%%%%%%%%%%%%%%%%%%%%%%%%%%%%%%%%%%%%%%%%%%%%%%%%%%%%%%%

\section{Related works}
\label{sec:related}

We position our work at the intersection of hallucination detection, representation engineering, and tractable probabilistic modeling.

\textbf{Hallucination detection}.
Early approaches exploit token probability and semantic entropy~\citep{farquhar_detecting_2024, azaria2023internal}, but suffer from LLM overconfidence and require multi-pass sampling. \citet{kossen2024semantic} reduce this cost via lightweight hidden-state probes, while HaloScope~\citep{du_haloscope_2024} extracts hallucination-prone features from unlabeled generations.
LLM-judge methods~\citep{friel_chainpoll_2023, heo_halucheck_2025} achieve strong precision at the cost of significant inference overhead. 
Our approach, instead, frames detection as exact density estimation over a contrastively trained latent manifold, yielding a principled, single-pass uncertainty signal.

\textbf{Hallucination correction}.
ROME~\citep{meng_locating_2022} pioneered weight editing for factual associations; ITI~\citep{li_inference-time_2023} and TruthX~\citep{zhang-etal-2024-truthx} shifted this to activation space.
Adaptive variants such as SADI~\citep{wang2025semanticsadaptiveactivationinterventionllms}, AdaSteer~\citep{zhao-etal-2025-adasteer}, and query-routed editing~\citep{liao_query-routed_2026} condition interventions on semantic context to mitigate over-editing, \textit{implicitly} recognizing the tension between correction and preservation that we \textit{formally quantify} as the \textit{Detection-Correction Asymmetry}.
Decoding-time methods, such as Contrastive Decoding~\citep{li-etal-2023-contrastive}, DoLa~\citep{chuang2024doladecodingcontrastinglayers}, and ICD~\citep{zhang-etal-2025-alleviating}, instead operate in token space, preserving fluency but lacking a principled uncertainty signal to determine \textit{when} to intervene.
\methodname inherits their manifold-safety property, while grounding intervention in exact latent density.

\textbf{Probabilistic Circuits}.
PCs guarantee exact marginal and Maximum A Posteriori (MAP) inference in linear time~\citep{poon2011spn, darwiche2003diff}, with extensions to online structure learning~\citep{hsu_online_2017}, lossless compression~\citep{liu_lossless_2022}, and neural integration~\citep{martires_probabilistic_2024, chen_neural_2025}. 
Their tractability is essential for our setting: hallucination gating requires per-token density evaluation at decoding time, where sampling-based estimators or implicit density models would impose prohibitive overhead. To our knowledge, we are the first to deploy PCs as inference-time density estimators over LLM residual streams.
%To our knowledge, we are the first to deploy PCs as inference-time density estimators over LLM residual streams.

\textbf{Geometry of LLM representations}.
Truthfulness has been shown to be linearly encoded in LLM hidden states~\citep{azaria2023internal, marks2024geometrytruthemergentlinear}, motivating latent-space hallucination control. 
Recent work confirms that outlier detection over representations can reduce hallucinations in downstream tasks~\citep{xie2026reducinghallucinationsllmbasedscientific, pan2026conversationalcontextclassificationrepresentation}. Our framework builds on this geometric intuition, replacing linear probes with a full probabilistic density model to capture the complex, heavy-tailed geometry of LLM representation spaces.

%%%%%%%%%%%%%%%%%%%%%%%%%%%%%%%%%%%%%%%%%%%%%%%%%%%%%%%%%%%%

\section{Methodology}
\label{sec:method}
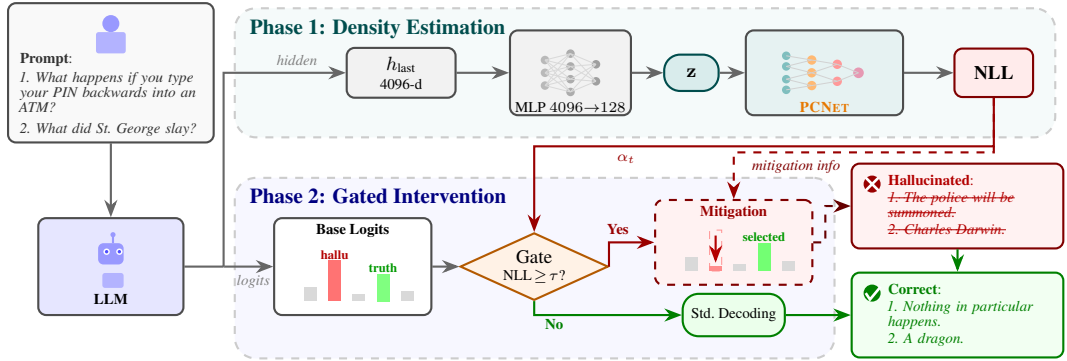
\begin{figure}[ht]
\centering
\begin{tikzpicture}[
    scale=0.8, transform shape, % Scaled to fit the wider horizontal layout
    >=Stealth, % Arrowhead style
    % --- GLOBAL STYLES ---
    base/.style={rectangle, rounded corners=3pt, draw=black!70, thick,
                 align=center, inner sep=4pt, font=\small},
    % --- INPUT & LLM STYLES ---
    input/.style={base, fill=gray!5, minimum width=3.4cm, minimum height=2.3cm,
                  align=left, inner sep=2pt},
    llm/.style={base, fill=blue!10, minimum width=2.4cm, minimum height=1.6cm},
    % --- PHASE 1 STYLES (Continuous/Density Estimation) ---
    embbox/.style={base, fill=gray!10, minimum width=1.8cm, minimum height=0.8cm},
    proj/.style={base, fill=gray!10, minimum width=2.0cm, minimum height=1.4cm},
    zlatent/.style={base, fill=teal!15, rounded corners=6pt,
                    minimum width=0.9cm, minimum height=0.6cm, draw=teal!60!black},
    pcnet/.style={base, fill=teal!10, minimum width=2.6cm, minimum height=1.4cm},
    nllbox/.style={base, fill=red!8, minimum width=1.3cm, minimum height=0.8cm,
                   draw=red!50!black},
    % --- PHASE 2 STYLES (Discrete/Intervention) ---
    logitbox/.style={rectangle, rounded corners=3pt, draw=black!70, thick,
                     fill=white, minimum width=2.4cm, minimum height=1.1cm},
    gate/.style={diamond, draw=orange!70!black, thick, fill=orange!10,
                 align=center, aspect=2.2, inner sep=1pt, font=\small},
    corrbox/.style={rectangle, rounded corners=3pt, draw=red!50!black, thick, dashed,
                    fill=red!5, minimum width=2.4cm, minimum height=1.1cm},
    safebox/.style={base, fill=green!8, rounded corners=6pt,
                    minimum width=1.6cm, minimum height=0.7cm, draw=green!50!black},
    % --- OUTPUT STYLES ---
    hallu/.style={rectangle, rounded corners=3pt, draw=red!50!black, thick,
                  fill=red!5, align=left, inner sep=5pt, font=\scriptsize,
                  minimum width=3.5cm, minimum height=1.4cm},
    factual/.style={rectangle, rounded corners=3pt, draw=green!50!black, thick,
                    fill=green!5, align=left, inner sep=5pt, font=\scriptsize,
                    minimum width=3.5cm, minimum height=1.4cm},
    % --- BACKGROUND & ARROW STYLES ---
    phase/.style={rectangle, rounded corners=8pt, draw=gray!40, dashed, thick,
                  inner sep=8pt},
    arr/.style={->, thick, black!60},
    redarr/.style={->, thick, red!60!black},
    greenarr/.style={->, thick, green!50!black},
]

% ============================================================
% 1. INPUT BLOCK
% ============================================================
\node[input] (inp) at (0, 0) {};

% --- Custom User Icon (Centered at the top) ---
\fill[blue!40] ([yshift=-10pt]inp.north) circle (4pt);
\fill[blue!30, rounded corners=1pt]
    ([xshift=-7pt, yshift=-14pt]inp.north) rectangle ++(14pt, -10pt);

\node[anchor=north, font=\scriptsize, text=black!80, text width=3.0cm]
    at ([yshift=-20pt]inp.north) {%
    \textbf{Prompt}:\\[2pt]
    {\fontsize{7}{7}\selectfont\itshape
    1.\ What happens if you type your PIN backwards into an ATM?\\[2pt]
    2.\ What did St.\ George slay?}};

% ============================================================
% 2. LLM BLOCK
% ============================================================
\node[llm] (llm) at (0, -3.2) {};

% --- Custom Robot/Bot Icon ---
\fill[blue!40, rounded corners=2pt]
    ([xshift=-6pt, yshift=1pt]llm.center) rectangle ++(12pt, 10pt);
\fill[white] ([xshift=-2pt, yshift=7pt]llm.center) circle (1.5pt);
\fill[blue!70] ([xshift=-2pt, yshift=7pt]llm.center) circle (0.8pt);
\fill[white] ([xshift=4pt, yshift=7pt]llm.center) circle (1.5pt);
\fill[blue!70] ([xshift=4pt, yshift=7pt]llm.center) circle (0.8pt);
\draw[blue!40, line width=0.8pt] ([xshift=1pt, yshift=11pt]llm.center) -- ++(0, 4pt);
\fill[blue!40] ([xshift=1pt, yshift=15pt]llm.center) circle (1.2pt);
\fill[blue!30, rounded corners=1pt]
    ([xshift=-4pt, yshift=-3pt]llm.center) rectangle ++(10pt, -7pt);

\node[font=\scriptsize\bfseries] at ([yshift=-15pt]llm.center) {LLM};
\draw[arr] (inp.south) -- (llm.north);

% ============================================================
% 3. PHASE 1: DENSITY ESTIMATION 
% ============================================================
% Shifted right to create more breathing room
\node[embbox] (emb) at (4.8, 0) {$h_{\text{last}}$\\[-1pt]{\scriptsize 4096-d}};

% --- MLP Projector ---
\node[proj] (mlp) at (7.6, 0) {};
\fill[gray!55] ([xshift=-12pt, yshift=6pt]mlp.center) circle (2pt);
\fill[gray!55] ([xshift=-12pt, yshift=0pt]mlp.center) circle (2pt);
\fill[gray!55] ([xshift=-12pt, yshift=-6pt]mlp.center) circle (2pt);
\fill[gray!65] ([yshift=9pt]mlp.center) circle (2pt);
\fill[gray!65] ([yshift=3pt]mlp.center) circle (2pt);
\fill[gray!65] ([yshift=-3pt]mlp.center) circle (2pt);
\fill[gray!65] ([yshift=-9pt]mlp.center) circle (2pt);
\fill[gray!75] ([xshift=12pt, yshift=4pt]mlp.center) circle (2pt);
\fill[gray!75] ([xshift=12pt, yshift=-4pt]mlp.center) circle (2pt);
\foreach \iy in {6,0,-6} { \foreach \hy in {9,3,-3,-9} {
    \draw[gray!30, line width=0.2pt] ([xshift=-12pt, yshift=\iy pt]mlp.center) -- ([yshift=\hy pt]mlp.center); }}
\foreach \hy in {9,3,-3,-9} { \foreach \oy in {4,-4} {
    \draw[gray!30, line width=0.2pt] ([yshift=\hy pt]mlp.center) -- ([xshift=12pt, yshift=\oy pt]mlp.center); }}
\node[font=\scriptsize] at ([yshift=-16pt]mlp.center) {MLP $4096{\to}128$};

% Latent space (z)
\node[zlatent] (z) at (9.6, 0) {$\mathbf{z}$};

% --- PCNet Graph ---
\node[pcnet] (pc) at (11.8, 0) {};
\fill[teal!50] ([xshift=-18pt, yshift=8pt]pc.center) circle (2pt);
\fill[teal!50] ([xshift=-18pt, yshift=3pt]pc.center) circle (2pt);
\fill[teal!50] ([xshift=-18pt, yshift=-3pt]pc.center) circle (2pt);
\fill[teal!50] ([xshift=-18pt, yshift=-8pt]pc.center) circle (2pt);
\fill[orange!55] ([xshift=-6pt, yshift=6pt]pc.center) circle (2.2pt);
\fill[orange!55] ([xshift=-6pt, yshift=0pt]pc.center) circle (2.2pt);
\fill[orange!55] ([xshift=-6pt, yshift=-6pt]pc.center) circle (2.2pt);
\fill[red!40] ([xshift=6pt, yshift=4pt]pc.center) circle (2.2pt);
\fill[red!40] ([xshift=6pt, yshift=-4pt]pc.center) circle (2.2pt);
\fill[purple!40] ([xshift=16pt]pc.center) circle (2.5pt);
\draw[gray!35, line width=0.2pt] ([xshift=-18pt, yshift=8pt]pc.center) -- ([xshift=-6pt, yshift=6pt]pc.center);
\draw[gray!35, line width=0.2pt] ([xshift=-18pt, yshift=3pt]pc.center) -- ([xshift=-6pt, yshift=6pt]pc.center);
\draw[gray!35, line width=0.2pt] ([xshift=-18pt, yshift=3pt]pc.center) -- ([xshift=-6pt, yshift=0pt]pc.center);
\draw[gray!35, line width=0.2pt] ([xshift=-18pt, yshift=-3pt]pc.center) -- ([xshift=-6pt, yshift=0pt]pc.center);
\draw[gray!35, line width=0.2pt] ([xshift=-18pt, yshift=-3pt]pc.center) -- ([xshift=-6pt, yshift=-6pt]pc.center);
\draw[gray!35, line width=0.2pt] ([xshift=-18pt, yshift=-8pt]pc.center) -- ([xshift=-6pt, yshift=-6pt]pc.center);
\draw[gray!35, line width=0.2pt] ([xshift=-6pt, yshift=6pt]pc.center) -- ([xshift=6pt, yshift=4pt]pc.center);
\draw[gray!35, line width=0.2pt] ([xshift=-6pt, yshift=0pt]pc.center) -- ([xshift=6pt, yshift=4pt]pc.center);
\draw[gray!35, line width=0.2pt] ([xshift=-6pt, yshift=0pt]pc.center) -- ([xshift=6pt, yshift=-4pt]pc.center);
\draw[gray!35, line width=0.2pt] ([xshift=-6pt, yshift=-6pt]pc.center) -- ([xshift=6pt, yshift=-4pt]pc.center);
\draw[gray!35, line width=0.2pt] ([xshift=6pt, yshift=4pt]pc.center) -- ([xshift=16pt]pc.center);
\draw[gray!35, line width=0.2pt] ([xshift=6pt, yshift=-4pt]pc.center) -- ([xshift=16pt]pc.center);
\node[font=\scriptsize\bfseries] at ([yshift=-16pt]pc.center) {\pcnet};

% Negative Log-Likelihood Output 
\node[nllbox] (nll) at (14.6, 0) {\textbf{NLL}};

% --- Internal Phase 1 flow ---
\draw[arr] (emb) -- (mlp);
\draw[arr] (mlp) -- (z);
\draw[arr] (z) -- (pc);
\draw[arr] (pc) -- (nll);

% ============================================================
% 4. PHASE 2: GATED INTERVENTION
% ============================================================
% Tightened gap (Moved from Y=-4.8 up to Y=-2.8 and -2.2)

% --- Base Logits Container ---
\node[logitbox, minimum width=2.6cm, minimum height=1.6cm] (logits) at (4.0, -3.2) {};
\node[anchor=north, font=\scriptsize\bfseries, inner sep=3pt] 
    (logitslbl) at (logits.north) {Base Logits};

\begin{scope}[shift={([yshift=12pt]logits.south)}, y=0.45cm, x=0.20cm]
    \fill[gray!30]  (-4,-0.4) rectangle (-3, 0.1); 
    \fill[red!55]   (-2,-0.4) rectangle (-1, 1.1); % Hallu
    \fill[gray!30]  ( 0,-0.4) rectangle ( 1,-0.15); 
    \fill[green!50] ( 2,-0.4) rectangle ( 3, 0.6);  % Truth
    \fill[gray!30]  ( 4,-0.4) rectangle ( 5,-0.05); 
    \node[font=\tiny\bfseries, text=red!70!black]  at (-1.5, 1.3) {hallu};
    \node[font=\tiny\bfseries, text=green!60!black] at ( 2.5, 0.85) {truth};
\end{scope}

% --- Decision Gate ---
\node[gate] (gate) at (7, -3.2) {Gate\\[-2pt]{\scriptsize $\text{NLL}\!\geq\!\tau$?}};

% --- Correction Action (Triggered on YES) ---
\node[corrbox, minimum width=2.6cm, minimum height=1.4cm] (corr) at (10.3, -2.8) {};
\node[anchor=north, font=\scriptsize\bfseries, text=red!60!black, inner sep=3pt]
    (corrlbl) at (corr.north) {Mitigation};

\begin{scope}[shift={([yshift=12pt]corr.south)}, y=0.45cm, x=0.20cm] 
    \fill[gray!30]  (-4,-0.4) rectangle (-3, 0.1);
    \draw[red!40, dashed] (-2,-0.4) rectangle (-1, 1.1); % Ghost penalty
    \fill[red!45]   (-2,-0.4) rectangle (-1,-0.25);      
    \draw[->, red!70!black, thick] (-1.5, 0.9) -- (-1.5, -0.1); 
    \fill[gray!30]  ( 0,-0.4) rectangle ( 1,-0.15);
    \fill[green!55] ( 2,-0.4) rectangle ( 3, 0.6);        % Argmax
    \fill[gray!30]  ( 4,-0.4) rectangle ( 5,-0.05);
    \node[font=\tiny\bfseries, text=green!60!black] at (2.5, 0.85) {selected};
\end{scope}

% --- Safe Action (Triggered on NO) ---
\node[safebox] (safe) at (10.3, -4) {{\scriptsize Std.\ Decoding}};

% ============================================================
% 5. OUTPUT EXAMPLES BLOCK (Completely OUTSIDE Phase 2)
% ============================================================
% Shifted far right to X=16.5, clear of all backgrounds
\node[hallu] (hallubox) at (14, -2.2) {}; 
\fill[red!60!black] ([xshift=10pt, yshift=-10pt]hallubox.north west) circle (4pt);
\draw[white, line width=1pt] ([xshift=7pt, yshift=-7pt]hallubox.north west) -- ++(6pt, -6pt);
\draw[white, line width=1pt] ([xshift=13pt, yshift=-7pt]hallubox.north west) -- ++(-6pt, -6pt);
\node[anchor=north west, font=\scriptsize, text=red!70!black, text width=3.4cm]
    at ([xshift=14pt, yshift=-2pt]hallubox.north west) {%
    \textbf{Hallucinated}:\\[1pt]
    {\fontsize{7}{7}\selectfont\itshape
    \sout{1.\ The police will be \\summoned.}\\
    \sout{2.\ Charles Darwin.}}};

\node[factual] (factbox) at (14, -4) {}; 
\fill[green!50!black] ([xshift=10pt, yshift=-10pt]factbox.north west) circle (4pt);
\draw[white, line width=1pt] ([xshift=7pt, yshift=-10pt]factbox.north west)
    -- ++(2pt, -3pt) -- ++(5pt, 6pt);
\node[anchor=north west, font=\scriptsize, text=green!50!black, text width=3.4cm]
    at ([xshift=14pt, yshift=-2pt]factbox.north west) {%
    \textbf{Correct}:\\[1pt]
    {\fontsize{7}{7}\selectfont\itshape
    1.\ Nothing in particular \\happens.\\
    2.\ A dragon.}};

% ============================================================
% 6. CROSS-PHASE CONTROL ARROWS & ROUTING
% ============================================================
% --- Hidden/Logits Bus Routing from LLM ---
% Extending the bus out slightly further to match the right-shifted boxes
\draw[thick, black!60] (llm.east) -- ++(0.65, 0) coordinate (split);
\draw[arr] (split) |- node[pos=0.8, above, font=\scriptsize\itshape, text=gray]{hidden} (emb.west);
\draw[arr] (split) |- node[pos=0.8, below, font=\scriptsize\itshape, text=gray]{logits} (logits.west);

% --- Phase 2 Routing ---
\draw[arr] (logits.east) -- (gate.west);
\draw[redarr] (gate.east) |- node[pos=0.6, above, font=\scriptsize\bfseries, text=red!60!black]{Yes} (corr.west);
\draw[greenarr] (gate.south) |- node[near start, below right=2pt,
    font=\scriptsize\bfseries, text=green!50!black]{No} (safe.west);

% --- Control Signals (Alpha and Nabla h) ---
% Tightly navigating the 1cm gap between Phase 1 and Phase 2
\draw[redarr] (nll.south) -- ++(0, -0.8) -| node[pos=0.4, below, font=\scriptsize, text=red!60!black]{$\alpha_t$} (gate.north);
\draw[redarr, dashed, sharp corners] (nll.south) -- ++(0, -0.9) -| node[pos=0.38, below, font=\scriptsize\itshape, text=red!50!black]{mitigation info} (corr.north);

% --- Final Output Arrows ---
\draw[->, dashed, red!40!black, thick] (corr.east) -- ++(0.2, 0) |- (hallubox.west);
\draw[greenarr] (safe.east) -- (factbox.west);
\draw[->, thick, green!50!black] (hallubox.south) -- (factbox.north);

% ============================================================
% 7. ALIGNED BACKGROUNDS (Outputs Excluded)
% ============================================================
% Explicitly set boundaries so p2bg tightly stops before the output boxes
\coordinate (p1_L) at (2.4, 0); \coordinate (p1_R) at (14, 0);
\coordinate (p2_L) at (2.4, -2.8); \coordinate (p2_R) at (11.6, -2.8); % Stops strictly at X=13.6

\begin{scope}[on background layer]
    \node[phase, fill=teal!4, fit=(p1_L)(emb)(mlp)(z)(pc)(nll)(p1_R)] (p1bg) {};
\end{scope}
\node[anchor=north west, font=\bfseries, text=teal!60!black]
    at ([xshift=4pt, yshift=-2pt]p1bg.north west) {Phase 1: Density Estimation};

\begin{scope}[on background layer]
    % Only fitting the internal Phase 2 boxes, NOT hallubox or factbox
    \node[phase, fill=blue!3, fit=(p2_L)(logitslbl)(logits)(gate)(safe)(corrlbl)(corr)(p2_R)] (p2bg) {};
\end{scope}
\node[anchor=north west, font=\bfseries, text=blue!50!black]
    at ([xshift=4pt, yshift=-2pt]p2bg.north west) {Phase 2: Gated Intervention};

\end{tikzpicture}
\caption{Architecture of the proposed framework. \textcolor{teal!60!black}{\textbf{Phase~1}} (top) projects $h_{\text{last}}$ through a Multi-Layer Perceptron (MLP) bottleneck into \pcnet for exact NLL computation. \textcolor{blue!50!black}{\textbf{Phase~2}} (bottom) gates on $\text{NLL} \geq \tau$: anomalous states are detected, and the next token is selected via density-penalized lookahead, while factual states proceed through standard decoding. The example shown is based on \methodname applied to Qwen3-4B.}
\label{fig:framework}
\end{figure}

Our framework introduces a unified, tractable approach to both detecting and correcting LLM hallucinations. In \Cref{sec:pc}, we formalize the construction of \pcnet, which acts as a density estimator over the LLM residual stream. In \Cref{sec:halu_dec}, we detail how the exact NLL derived from \pcnet serves as a highly accurate anomaly detector. Finally, in \Cref{sec:halu_correction}, we introduce \methodname, a manifold-preserving decoding strategies that rely on \pcnet to steer the generation process. \Cref{fig:framework} shows the overall framework workflow. Theoretical guarantees (consistency, no-regret on confident states, and complexity analysis) are deferred to \Cref{app:theoretical}.

\subsection{Tractable density estimation via \pcnet}
\label{sec:pc}

To establish a mathematically rigorous safeguard over the LLM's continuous latent space, we design a tractable density estimator based on PCs~\citep{martires_probabilistic_2024, chen_neural_2025}. PCs are a class of deep generative models formulated as parameterized Directed Acyclic Graphs (DAGs) that guarantee exact and efficient marginal and conditional inference~\citep{liu_lossless_2022}.

Our architecture, referred to as \pcnet, is based on Neural Probabilistic Circuits~\citep{chen_neural_2025}, and is integrated directly into the final transformer block of the LLM. 
Let $h \in \mathbb{R}^{D_{LLM}}$ denote the activation of the residual stream immediately preceding the final embedding layer. 
To mitigate the curse of dimensionality and filter out syntactic noise, we first project $h$ through an information bottleneck $f_\phi : \mathbb{R}^{D_{LLM}} \to \mathbb{R}^{D_{PC}}$ to yield a low-dimensional representation $z \in \mathbb{R}^{D_{PC}}$ (where $D_{PC} \ll D_{LLM}$). 
The \pcnet then models the joint probability distribution $P(z)$ over this compressed latent space.

\paragraph{Graph topology and node formulation}
The \pcnet is structured as a layered DAG consisting of three fundamental node types: Input Nodes, Sum Nodes, and Product Nodes.
In the following, we first establish the standard definition of a PC, and then we detail our specific architectural modifications.

\begin{definition}[Probabilistic Circuit]\label{def:pc}
A Probabilistic Circuit over variables $Z = (Z_1,\dots,Z_m)$ is a rooted DAG $\mathcal{C}=(\mathcal{N},\mathcal{E})$ in which every node $n\in\mathcal{N}$ has a scope $\mathrm{sc}(n) \subseteq \{1,\dots,m\}$ and computes a non-negative function $\mathcal{C}_n : \mathrm{dom}(Z_{\mathrm{sc}(n)}) \to \mathbb{R}_{\ge 0}$ defined as:
\begin{align*}
\text{input (leaf)}:\quad & \mathcal{C}_n(z_{\mathrm{sc}(n)}) \;=\; q_n(z_{\mathrm{sc}(n)}; \eta_n), \\
\text{sum node}:\quad & \mathcal{C}_n(z_{\mathrm{sc}(n)}) \;=\; \sum_{c\in\mathrm{ch}(n)} w_{n,c}\,\mathcal{C}_c(z_{\mathrm{sc}(c)}),\quad w_{n,c}\ge 0,\quad \sum_c w_{n,c}=1, \\
\text{product node}:\quad & \mathcal{C}_n(z_{\mathrm{sc}(n)}) \;=\; \prod_{c\in\mathrm{ch}(n)} \mathcal{C}_c(z_{\mathrm{sc}(c)}),
\end{align*}
where $q_n(\,\cdot\,; \eta_n)$ is a tractable parametric density with learnable parameters $\eta_n$, and $\mathrm{ch}(n)$ denotes the children of $n$. We assume $\mathcal{C}$ is smooth and decomposable, so that $\mathcal{C}(z) := \mathcal{C}_{\text{root}}(z)$ is a valid density on $Z$ admitting exact, linear-time inference in $|\mathcal{N}|$~\citep{darwiche2003diff,poon2011spn}.
\end{definition}

Building upon this foundational structure, we tailor the internal node operations to capture the specific geometric properties of LLM representations as follows.

\textbf{Product and sum nodes}. 
In our architecture, \textit{Product Nodes} encode context-specific independence assumptions among disjoint subsets of the latent features. Conversely, \textit{Sum Nodes} model distinct latent sub-populations, such as divergent semantic or factual manifolds, by computing a convex combination of their children.

\textbf{Input nodes (heterogeneous mixture leaves)}.
Unlike traditional PCs relying on simple, single-distribution leaf nodes~\citep{hsu_online_2017}, the continuous latent representations of LLMs are known to exhibit complex, heavy-tailed geometries. Standard isolated parametric distributions often fail to capture these accurately. 
To address this, we introduce a mixed continuous distribution at each leaf. 

For each feature dimension $z_i$, the corresponding \textit{Input Node} computes the log-likelihood as a learned mixture of Gaussian ($G$), Laplace ($L$), and Student-T ($T$) distributions:
\begin{equation}
    \log P(z_i) = \sigma(g) \cdot \log \sum_{k \in \{G, L, T\}} w_k \exp(\log P_k(z_i \mid \mu, s, \nu))
\end{equation}
where: $g$ is a dimension-specific, learnable gating parameter, whose purpose is to dynamically scale the overall log-likelihood contribution of the $i$-th feature to the broader circuit; $\mu$ is a shared location parameter across all three distributions; $s$ is a shared scale parameter; and $\nu$ is the degrees of freedom parameter, specific to the Student-T component.

\textbf{Hierarchical construction and the classifier root}.
The \pcnet is initialized via a randomized, bottom-up construction algorithm. Following the instantiation of $D_{PC}$ leaf nodes, the network is built hierarchically by alternating product and sum layers up to a maximum depth $L_{PC}$. At depth $L_{PC}$, the DAG terminates in a root node, $\mathcal{C}_{\text{root}}$, which computes the unnormalized joint log-probability of the input $z$, yielding a single scalar density estimate used directly as the anomaly score.

\subsection{Latent anomaly detection via exact marginals}
\label{sec:halu_dec}

\begin{figure}
    \centering
    \includegraphics[width=\linewidth]{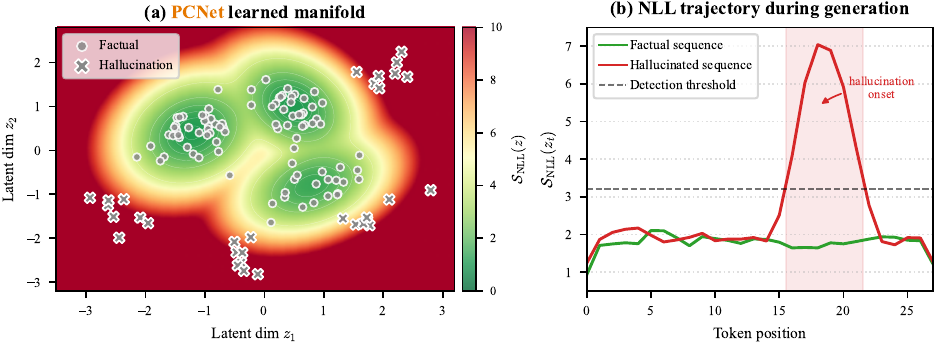}
    \caption{Illustration of the \pcnet density model.
    \textbf{(a)} Factual hidden-state projections concentrate in high-density regions of the learned manifold; hallucinated projections fall into low-density outlier regions where $\mathcal{S}_\text{NLL}$ is elevated. \textbf{(b)} Per-token NLL trajectory: factual generation remains stable while a hallucination triggers a sharp spike that crosses the detection threshold.}
    \label{fig:z_distribution}
    \vspace{-0.5cm}
\end{figure}

We integrate the \pcnet into the standard LLM inference pipeline to detect hallucinations. 
% To ensure our framework scales to state-of-the-art architectures without prohibitive memory overhead, the target LLM is frozen and loaded using 4-bit NormalFloat (NF4) quantization~\citep{dettmers2023qlora7}. 
Given an input prompt $x$, we execute a standard forward pass and isolate the final-layer hidden state corresponding strictly to the last valid token, $h \in \mathbb{R}^{D_{LLM}}$. Because the raw residual stream is high-dimensional (e.g., $D_{LLM} = 4096$), we employ the information bottleneck (a 2-layer MLP with ReLU activations) $f_\phi$:
\begin{equation}
    z = f_\phi(h) \in \mathbb{R}^{D_{PC}}.
\end{equation}
This projection compresses the hidden state into a compact continuous vector $z$, filtering out syntactic artifacts while preserving the core semantic and factual geometries~\citep{azaria2023internal}. 
During deterministic generation, the projected state $z$ is continuously evaluated by the \pcnet. 
By exploiting the circuit's guarantee of exact marginal inference, we compute the NLL of the current state to serve as our uncertainty metric:
\begin{equation}
  \mathcal{S}_{\text{NLL}}(z) = - \log \mathcal{C}_{\text{root}}(z).
\end{equation}
Following recent literature on hidden state forensics~\citep{xie2026reducinghallucinationsllmbasedscientific}, we operationalize factually grounded generations as residing within high-density regions of this learned manifold. Consequently, when an LLM trajectory deviates into a hallucination, its projection falls into a low-density region, triggering a sharp spike in $\mathcal{S}_{\text{NLL}}(z)$, as shown in \Cref{fig:z_distribution}.

%This rigorous uncertainty quantification serves as an exact geometric anomaly detector, forming the mathematical foundation for the dynamic decoding interventions detailed in subsequent sections.

\paragraph{Contrastive manifold optimization}
To learn the manifold of truthful representations, we jointly optimize the \pcnet parameters $\theta$ and the projection bottleneck $f_\phi$, while the base LLM remains frozen. Given a dataset of paired factual ($h^+$) and hallucinated ($h^-$) hidden states, we compute their corresponding projections $z^+ = f_\phi(h^+)$ and $z^- = f_\phi(h^-)$. 
We then minimize a composite objective that balances generative density estimation with a contrastive margin penalty:
\begin{equation}
    \mathcal{L}(\theta, \phi) = \alpha \underbrace{\mathbb{E}_{h^+}[-\log \mathcal{C}_{\text{root}}(z^+)]}_{\text{Generative NLL}} + (1 - \alpha) \underbrace{\mathbb{E}_{h^+, h^-}[\max(0, \gamma + \log \mathcal{C}_{\text{root}}(z^-) - \log \mathcal{C}_{\text{root}}(z^+))]}_{\text{Contrastive Margin}}
\end{equation}
where $\gamma > 0$ defines the geometric margin and $\alpha \in [0, 1]$ controls the loss weighting. 
The generative term encourages the \pcnet to construct a high-density manifold encompassing factually grounded states. Concurrently, the contrastive term enforces a strict geometric separation, actively displacing hallucinated states into low-density regions bounded by the margin $\gamma$. The joint parameters $\{\theta, \phi\}$ are optimized end-to-end via Adam~\citep{kingma2017adammethodstochasticoptimization}. To prevent numerical instability during the exact marginal computations across the circuit, we apply gradient clipping ($\|\nabla\|_2 \le 1.0$). We establish consistency of this estimator under standard regularity conditions in \Cref{app:consistency}.

\subsection{Manifold-preserving hallucination correction via \methodname}
\label{sec:halu_correction}

Having established the \pcnet as a tractable density estimator capable of identifying out-of-distribution latent states, we address the Detection-Correction Asymmetry. Moreover, we introduce Probabilistic Circuits Latent Density Contrastive Decoding (\methodname), departing from unconstrained continuous latent changes of the LLMs, which contribute to semantic collapse. \methodname, instead, is governed by a dynamic gating mechanism, intervening only when the geometry of the hidden state deviates from the factual manifold.

\paragraph{Dynamic intervention gating}
At each decoding timestep $t$, the LLM produces a final hidden state $h_t \in \mathbb{R}^{D_{LLM}}$. We project $h_t$ into the low-dimensional semantic space $z_t = f_\phi(h_t)$ and compute its exact NLL, $\mathcal{S}_{\text{NLL}}(z_t) = -\log \mathcal{C}_{\text{root}}(z_t)$. To route compute efficiently, we define a dynamic intervention strength, $\beta_t \in (0,1)$, using a sigmoid centered around a calibrated anomaly threshold $\tau$. The threshold $\tau$ is based on a validation set of observed NLL scores, determined by calculating the precision-recall curve against ground truth labels and selecting the threshold that yields the maximum F1 score $\beta_t = \sigma\left(\mathcal{S}_{\text{NLL}}(z_t) - \tau\right)$. If $\beta_t$ falls below a conservative margin (e.g., $\beta_t < 0.05$), the state is deemed geometrically stable. In this case, the system bypasses all interventions, defaulting to standard $\mathcal{O}(1)$ decoding (see \Cref{app:complexity}). This gating provably preserves the base LLM's behavior on high-confidence factual states (\Cref{prop:gate-regret}).

\paragraph{PC-latent density contrastive decoding (\methodname)}
When an active anomaly is detected, \methodname suspends standard decoding to perform a non-destructive, $k$-candidate lookahead search in the discrete token space. We extract the top-$k$ most probable tokens from the raw logits, $\{c_1, \dots, c_k\}$, and compute the hypothetical future latent states, $h_{t+1}^{(c_i)}$.
The optimal token $x_t$ is selected by maximizing a penalized log-probability:
\begin{equation}\label{eq:score}
\text{Score}_{\text{LDCD}}(c_i) = \log P_{\text{LM}}(c_i | x_{<t}) - \beta_t \cdot \mathcal{S}_{\text{NLL}}(f_\phi(h_{t+1}^{(c_i)})).
\end{equation}
\Cref{app:complexity} demonstrates the computational overhead of the method. This formulation fundamentally advances heuristic contrastive methods~\citep{li-etal-2023-contrastive}. Instead of penalizing generations using an ``amateur'' proxy model, \methodname contrasts generative confidence directly against the exact log-density score under the learned manifold model.
Because $\beta_t$ dynamically scales this penalty, the model generates fluent syntax when safe, but is constrained from selecting tokens that deepen a hallucinated trajectory.

%%%%%%%%%%%%%%%%%%%%%%%%%%%%%%%%%%%%%%%%%%%%%%%%%%%%%%%%%%%%

\section{Experiments}
\label{sec:exp_eval}
We outline the key design choices and evaluation protocol below. Full details on hardware, software dependencies, RNG seeds, and reproducibility, as well as the used LLMs, benchmark datasets, and evaluation metrics are provided in \Cref{app:exp_set}. We develop our experimental evaluation in three main research questions:

\textbf{RQ1} How effective is tractable density estimation via \pcnet at identifying hallucinations in LLMs?\\
\textbf{RQ2} How do gating interventions using PC-derived uncertainty (\pcnet) mitigate the Detection-Correction Asymmetry (\methodname)?\\
\textbf{RQ3} How does our \methodname compare to state-of-the-art interventions in terms of correction efficacy and text fluency?

\textbf{Baselines}. We evaluate \pcnet against four established uncertainty estimators: \textbf{Token-NLL}, a standard output-level heuristic based on softmax probabilities~\citep{azaria2023internal}; \textbf{SEP} (Semantic Entropy Probes)~\citep{kossen2024semantic}, which trains lightweight linear probes on hidden states to predict generative semantic variance; \textbf{Haloscope}~\citep{du_haloscope_2024}, which utilizes specialized latent feature extraction to identify ungrounded representations; and \textbf{AutoFact}~\citep{heo_halucheck_2025}, an NLI-based judge adapted by replacing external document retrieval with the in-context passage as the NLI premise, since our QA datasets provide grounding context.

For the mitigation stage, we evaluate \methodname against five established hallucination reduction methods and a \textbf{Vanilla} baseline (no intervention).
We consider \textbf{DoLa}~\citep{chuang2024doladecodingcontrastinglayers}, which contrasts logit distributions between mature and premature layers; \textbf{ITI}~\citep{li_inference-time_2023}, which identifies truthful directions via linear probing of attention heads; \textbf{AdaSteer}~\citep{zhao-etal-2025-adasteer} and \textbf{SADI}~\citep{wang2025semanticsadaptiveactivationinterventionllms}, two adaptive activation steering methods that condition interventions on semantic context; and \textbf{ICD}~\citep{zhang-etal-2025-alleviating}, which constructs a contrastive signal from deliberately induced hallucinations.

\paragraph{RQ1: efficacy of \pcnet for latent hallucinations detection}

%\begin{figure}
%    \centering
%    \includegraphics[width=\textwidth]{plots/detection_performance.pdf}
%    \caption{Unified hallucination detection performance. \textbf{(a)} Average AUROC and \textbf{(b)} F1 Score demonstrate that \pcnet consistently outperforms likelihood and entropy-based baselines across all generative contexts, achieving near-perfect separation on \textsc{CoQA} and \textsc{SQuAD}. \textbf{(c)} Scalability heatmap confirms \pcnet’s robustness across diverse model architectures and sizes, maintaining high reliability even on adversarial benchmarks like \textsc{TruthfulQA}.}
%    \label{fig:detection}
%\end{figure}

\begin{table}[!ht]
\centering
\caption{Hallucination detection performance. We report AUROC and F1 across four datasets and three seeds (mean\std{std}). Best results per model and dataset are highlighted in \textbf{bold}. The \textbf{Avg.} column is averaged across all datasets and seeds.}
\label{tab:detection_performance}
\resizebox{\textwidth}{!}{
\small
\begin{tabular}{ll cc cc cc cc cc}
\toprule
\multirow{2}[2]{*}{\textbf{Model}} & \multirow{2}[2]{*}{\textbf{Method}}
  & \multicolumn{2}{c}{\textbf{CoQA}}
  & \multicolumn{2}{c}{\textbf{SQuAD v2.0}}
  & \multicolumn{2}{c}{\textbf{TriviaQA}}
  & \multicolumn{2}{c}{\textbf{TruthfulQA}}
  & \multicolumn{2}{c}{\textbf{Avg.}} \\
\cmidrule(lr){3-4}\cmidrule(lr){5-6}\cmidrule(lr){7-8}\cmidrule(lr){9-10}\cmidrule(lr){11-12}
& & AUROC & F1 & AUROC & F1 & AUROC & F1 & AUROC & F1 & AUROC & F1 \\
\midrule
\multirow{5}{*}{\textbf{Llama-3.2-1B}}
& Token NLL
  & 0.55\std{.00} & 0.67\std{.00}
  & 0.61\std{.01} & 0.68\std{.01}
  & 0.76\std{.06} & 0.74\std{.02}
  & 0.47\std{.01} & 0.67\std{.00}
  & 0.59\std{.12} & 0.69\std{.03} \\
& SEP
  & 0.77\std{.07} & 0.77\std{.02}
  & 0.62\std{.09} & 0.67\std{.01}
  & 0.56\std{.03} & 0.67\std{.00}
  & 0.53\std{.03} & 0.67\std{.00}
  & 0.62\std{.11} & 0.70\std{.05} \\
& HaloScope
  & 0.85\std{.02} & 0.85\std{.03}
  & 0.95\std{.05} & 0.91\std{.06}
  & 0.86\std{.11} & 0.81\std{.10}
  & 0.61\std{.02} & 0.67\std{.00}
  & 0.82\std{.14} & 0.81\std{.11} \\
& AutoFact
  & 0.56\std{.02} & 0.67\std{.00}
  & 0.58\std{.01} & 0.68\std{.01}
  & 0.80\std{.02} & 0.76\std{.00}
  & 0.34\std{.01} & 0.67\std{.00}
  & 0.57\std{.17} & 0.69\std{.04} \\
\rowcolor{pcorangelight} \cellcolor{white} &
\pcnetplain \textbf{(Ours)}
  & \textbf{0.95\std{.05}} & \textbf{0.95\std{.03}}
  & \textbf{0.98\std{.01}} & \textbf{0.97\std{.02}}
  & \textbf{0.91\std{.04}} & \textbf{0.90\std{.07}}
  & \textbf{0.66\std{.10}} & \textbf{0.69\std{.03}}
  & \textbf{0.88\std{.14}} & \textbf{0.88\std{.12}} \\
\midrule
\multirow{5}{*}{\textbf{Mistral-7B-v0.3}}
& Token NLL
  & 0.56\std{.00} & 0.67\std{.00}
  & 0.62\std{.00} & 0.68\std{.02}
  & 0.82\std{.01} & 0.78\std{.00}
  & 0.52\std{.01} & 0.67\std{.00}
  & 0.63\std{.13} & 0.70\std{.05} \\
& SEP
  & 0.69\std{.12} & 0.72\std{.03}
  & 0.64\std{.07}  & 0.71\std{.01}
  & 0.63\std{.06} & 0.68\std{.01}
  & 0.54\std{.04} & 0.67\std{.01}
  & 0.62\std{.09} & 0.69\std{.03} \\
& HaloScope
  & 0.88\std{.03} & 0.88\std{.02}
  & 0.90\std{.05} & 0.90\std{.04}
  & 0.88\std{.06} & 0.87\std{.03}
  & 0.55\std{.05} & 0.67\std{.00}
  & 0.79\std{.16} & 0.82\std{.10} \\
& AutoFact
  & 0.56\std{.02} & 0.67\std{.00}
  & 0.58\std{.01} & 0.68\std{.01}
  & 0.80\std{.02} & 0.76\std{.00}
  & 0.34\std{.01} & 0.67\std{.00}
  & 0.57\std{.18} & 0.70\std{.04} \\
\rowcolor{pcorangelight} \cellcolor{white} &
\pcnetplain \textbf{(Ours)}
  & \textbf{0.97\std{.02}} & \textbf{0.96\std{.03}}
  & \textbf{0.98\std{.00}} & \textbf{0.97\std{.00}}
  & \textbf{0.98\std{.01}} & \textbf{0.96\std{.01}}
  & \textbf{0.79\std{.06}} & \textbf{0.76\std{.04}}
  & \textbf{0.92\std{.09}} & \textbf{0.91\std{.10}} \\
\midrule

\multirow{5}{*}{\textbf{Qwen3-4B}}
& Token NLL
  & 0.55\std{.00} & 0.67\std{.00}
  & 0.61\std{.01} & 0.69\std{.01}
  & 0.76\std{.01} & 0.74\std{.01}
  & 0.53\std{.01} & 0.67\std{.00}
  & 0.61\std{.09} & 0.69\std{.03} \\
& SEP
  & 0.65\std{.24} & 0.73\std{.07}
  & 0.42\std{.09}  & 0.67\std{.01}
  & 0.69\std{.02} & 0.72\std{.01}
  & 0.54\std{.08} & 0.67\std{.00}
  & 0.60\std{.15} & 0.70\std{.04} \\
& HaloScope
  & 0.63\std{.09} & 0.73\std{.05}
  & 0.82\std{.09} & 0.80\std{.07}
  & 0.72\std{.08} & 0.72\std{.04}
  & 0.53\std{.02} & 0.68\std{.01}
  & 0.67\std{.13} & 0.73\std{.06} \\
& AutoFact
  & 0.56\std{.02} & 0.67\std{.00}
  & 0.58\std{.01} & 0.68\std{.01}
  & 0.80\std{.02} & 0.76\std{.00}
  & 0.34\std{.01} & 0.67\std{.00}
  & 0.57\std{.17} & 0.70\std{.04} \\
\rowcolor{pcorangelight} \cellcolor{white} &
\pcnetplain \textbf{(Ours)}
  & \textbf{0.96\std{.06}} & \textbf{0.96\std{.06}}
  & \textbf{0.97\std{.01}} & \textbf{0.96\std{.01}}
  & \textbf{0.95\std{.04}} & \textbf{0.93\std{.03}}
  & \textbf{0.81\std{.02}} & \textbf{0.78\std{.02}}
  & \textbf{0.92\std{.07}} & \textbf{0.91\std{.08}} \\
\midrule

\multirow{5}{*}{\textbf{Llama-3.1-8B}}
& Token NLL
  & 0.55\std{.00} & 0.67\std{.00}
  & 0.61\std{.01} & 0.67\std{.00}
  & 0.82\std{.01} & 0.76\std{.01}
  & 0.49\std{.00} & 0.67\std{.00}
  & 0.62\std{.13} & 0.69\std{.04} \\
& SEP
  & 0.33\std{.20} & 0.67\std{.01}
  & 0.57\std{.09} & 0.67\std{.01}
  & 0.52\std{.09} & 0.67\std{.01}
  & 0.43\std{.01} & 0.67\std{.00}
  & 0.46\std{.14} & 0.67\std{.01} \\
& HaloScope
  & 0.97\std{.02} & 0.97\std{.02}
  & \textbf{0.97\std{.02}} & \textbf{0.96\std{.01}}
  & 0.95\std{.06} & 0.92\std{.03}
  & 0.59\std{.03} & 0.67\std{.00}
  & 0.87\std{.17} & 0.88\std{.13} \\
& AutoFact
  & 0.56\std{.01} & 0.67\std{.00}
  & 0.58\std{.01} & 0.67\std{.00}
  & 0.80\std{.01} & 0.76\std{.00}
  & 0.35\std{.00} & 0.67\std{.00}
  & 0.57\std{.17} & 0.69\std{.04} \\
\rowcolor{pcorangelight} \cellcolor{white} &
\pcnetplain \textbf{(Ours)}
  & \textbf{0.98\std{.02}} & \textbf{0.98\std{.01}}
  & \textbf{0.98\std{.02}} & \textbf{0.97\std{.02}}
  & \textbf{0.97\std{.02}} & \textbf{0.97\std{.01}}
  & \textbf{0.75\std{.06}} & \textbf{0.72\std{.04}}
  & \textbf{0.92\std{.11}} & \textbf{0.91\std{.12}} \\
\bottomrule
\end{tabular}
}
\end{table}

\Cref{tab:detection_performance} reports detection performance across all models and benchmarks. 
\pcnet consistently outperforms all baselines, achieving near-perfect separation on CoQA and SQuAD~v2.0, confirming that exact density estimation is a far more reliable hallucination detector than output-space heuristics or discriminative probes.
Token NLL and SEP perform near random guess across nearly all settings, consistent with known LLM overconfidence~\citep{azaria2023internal}. 
This degradation is particularly pronounced in smaller models, where the output distribution becomes overly diffuse, rendering token-level probability signals an unreliable proxy for internal factual grounding.
HaloScope is a stronger baseline on CoQA and SQuAD, but degrades significantly on knowledge-intensive and adversarial benchmarks, collapsing on TruthfulQA.
We attribute this to the absence of a principled density model, making it unable to generalize across different instruction-tuning regimes.
TruthfulQA remains the hardest setting for all methods, as its questions target misconceptions deeply encoded in pretraining weights~\citep{lin2022truthfulqameasuringmodelsmimic}. 
Yet, \pcnet leads across all models also on this dataset, with Mistral-7B achieving the highest AUROC, consistent with the hypothesis that more capable models develop more geometrically separable truth representations~\citep{marks2024geometrytruthemergentlinear}.
% Finally, detection performance is stable across all tested architectures and scales, from Llama3.2-1B to Llama3.1-8B, validating the transferability of our approach.

\paragraph{RQ2: mitigating the detection-correction symmetry}

\begin{wraptable}{r}{0.60\textwidth}
\centering
\caption{Gating advantage summary.
  \textbf{Corr} denotes the percentage of originally correct generations degraded by indiscriminate un-gated intervention. \textbf{Pres.} denotes the percentage of those correct generations successfully protected by \pcnet gating. (U-G) = Un-Gated, (G) = Gated. Results are computed across models and seeds (mean\std{std}).}
\label{tab:gate_summary}
\resizebox{0.59\textwidth}{!}{%
\setlength{\tabcolsep}{2pt}
\small
\begin{tabular}{lcccccccc}
\toprule
& \multicolumn{2}{c}{\textbf{SQuAD} $\uparrow$}
& \multicolumn{2}{c}{\textbf{T+I} $\uparrow$}
& \multicolumn{2}{c}{\textbf{TriviaQA} $\uparrow$}
& \textbf{Corr.$\downarrow$}
& \textbf{Pres.$\uparrow$} \\
\cmidrule(lr){2-3}\cmidrule(lr){4-5}\cmidrule(lr){6-7}
\textbf{Method} & U-G & G & U-G & G & U-G & G & \% & \% \\
\midrule
DoLa
  & 0.77\std{.07} & 0.86\std{.04}
  & 0.55\std{.12} & 0.63\std{.12}
  & 0.50\std{.16} & 0.72\std{.10}
  & 55.3\std{22.4} & 77.8\std{23.4} \\
ITI
  & 0.13\std{.24} & 0.56\std{.12}
  & 0.18\std{.17} & 0.42\std{.17}
  & 0.02\std{.03} & 0.50\std{.02}
  & 63.5\std{25.2} & 76.9\std{23.7} \\
AdaSteer
  & 0.83\std{.05} & \textbf{0.87\std{.03}}
  & 0.58\std{.10} & 0.66\std{.09}
  & 0.55\std{.15} & 0.75\std{.09}
  & 55.8\std{23.7} & 78.1\std{24.1} \\
SADI
  & 0.83\std{.06} & \textbf{0.87\std{.04}}
  & 0.58\std{.11} & 0.66\std{.10}
  & \textbf{0.56\std{.15}} & \textbf{0.75\std{.09}}
  & \textbf{53.7\std{24.1}} & 78.6\std{24.1} \\
ICD
  & 0.66\std{.17} & 0.81\std{.09}
  & 0.55\std{.09} & 0.64\std{.09}
  & 0.45\std{.13} & 0.70\std{.08}
  & 57.0\std{20.8} & 77.2\std{24.2} \\
\rowcolor{methodpurplelight}
\methodnameplain
  & 0.80\std{.06} & \textbf{0.87\std{.04}}
  & \textbf{0.62\std{.12}} & \textbf{0.67\std{.11}}
  & 0.49\std{.15} & 0.72\std{.09}
  & \textbf{53.7\std{25.2}} & \textbf{79.3\std{24.2}} \\
\midrule
\textit{Avg.}
  & 0.67\std{.28} & 0.81\std{.14}
  & 0.51\std{.19} & 0.61\std{.14}
  & 0.43\std{.23} & 0.69\std{.12}
  & 56.5\std{23.8} & 78.0\std{24.0} \\
\bottomrule
\end{tabular}}
\end{wraptable}

\begin{figure*}[t]
    \centering  
    \includegraphics[width=0.9\textwidth]{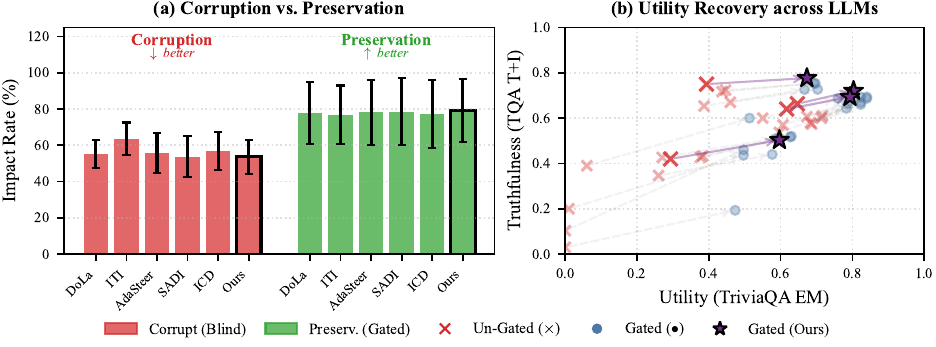}
    \caption{\textbf{(a)} Corruption (Red) and Preservation (Green) rates across all models and methods, averaged across LLMs with standard deviation error bars. \textit{Corruption} measures the fraction of correct generations degraded by un-gated intervention; 
    \textit{Preservation} measures those protected by \pcnet gating. \textbf{(b)} Utility-truthfulness trade-off across the four tested LLMs. Un-gated interventions (Red) trigger semantic collapse; \pcnet gating (Blue) recovers utility (SQuAD/Trivia EM) and shifts toward the optimal frontier. Each data point reports a separate experimental setting.}
    \label{fig:pareto_impact}
\end{figure*}

\Cref{tab:gate_summary} and \Cref{fig:pareto_impact} report the impact of \pcnet gating across all correction methods. The results provide strong empirical evidence for the Detection-Correction Asymmetry: applying corrections indiscriminately at every token consistently degrades performance. 
Gating via \pcnet avoids this collapse, recovering substantial performance across all benchmarks.
Our \methodname achieves the lowest mean corruption rate and the highest preservation rate across all methods, while matching the best baselines on SQuAD EM and leading on TQA T+I. 
This demonstrates that operating in the token space, guided by exact density estimation, is safer than editing continuous hidden states.

\paragraph{RQ3: evaluating correction strategies}
\Cref{tab:truthfulqa_results} and \Cref{tab:qa_benchmarks_results} report the  correction performance across all models and benchmarks.
On TruthfulQA, \methodname achieves the highest True+Info score on Qwen3-4B ($0.78$, $+0.28$) and Mistral-7B ($0.72$, $+0.22$), and ties for the best on Llama-3.1-8B. Across all models, \methodname consistently leads on MC2 and MC3 (the metrics measuring calibrated uncertainty over multiple truthful completions) with notable gains on  Mistral-7B (MC2 $0.72$, MC3 $0.29$), substantially outperforming all baselines. 
This pattern suggests that density-guided discrete decoding produces more robustly truthful output distributions rather than merely shifting single-answer confidence. 
\methodname does not lead on MC1, which measures hard single-answer accuracy against the most likely misconception. 
This is expected: MC1 rewards aggressive logit re-ranking toward a single correct answer, a behavior better suited to contrastive methods such as DoLa and ICD that directly amplify output probability gaps. Moreover, recent analyses have shown that MC1 itself has structural limitations, with up to $25\%$ of samples scored as incorrect potentially being factually correct due to log-probability rankings being sensitive to generation tone rather than factual accuracy~\citep{bang2025hallulens}.
\methodname instead operates by penalizing tokens that deepen latent anomalies, naturally producing broader distributional corrections that benefit MC2 and MC3 but do not sharply concentrate mass on a single choice. 

On standard QA benchmarks, see the full results shown in \Cref{app:res_tables} (\Cref{tab:qa_benchmarks_results}), \methodname consistently matches the best-performing baselines while remaining competitive across all models, achieving SQuAD EM $0.87$ and CoQA F1 $0.74$ on Mistral-7B, and SQuAD EM $0.85$ on Llama-3.2-1B, trailing the best baseline by at most $0.01$. 
Furthermore, the IGR varies meaningfully across models ($41\%$--$77\%$), confirming that \pcnet intervenes selectively rather than blindly correcting every token.
% Table 1: TruthfulQA Results
% ============================================================
\begin{table}[t]
\centering
\vspace{-0.6cm}
\caption{Performance on TruthfulQA (gated mode). Results (mean\std{std} across three seeds) are formatted as \textit{score}\std{std} ($\Delta$\std{std}). IGR (\%) reflects the fraction of prompts where \pcnet actively triggered an intervention. Best post-intervention scores per model are highlighted in \textbf{bold}.}
\label{tab:truthfulqa_results}
\resizebox{\textwidth}{!}{
\small
\begin{tabular}{llccccc}
\toprule
\textbf{Model} & \textbf{Method} & \textbf{IGR (\%)}
& \textbf{T+I} $\uparrow$ & \textbf{MC1} $\uparrow$
& \textbf{MC2} $\uparrow$ & \textbf{MC3} $\uparrow$ \\
\midrule

\multirow{7}{*}{\textbf{Llama-3.2-1B}}
& Vanilla   & 0.0
  & 0.50\std{.00} & 0.23\std{.02} & 0.41\std{.01} & 0.06\std{.00} \\
& DoLa      & 77.9\std{15.4}
  & 0.44\std{.07} (-0.06\std{.07})
  & \textbf{0.54\std{.11} (+0.31\std{.09})}
  & 0.26\std{.05} (-0.15\std{.07})
  & 0.04\std{.02} (-0.02\std{.03}) \\
& ITI       & 77.9\std{15.4}
  & 0.19\std{.11} (-0.31\std{.11})
  & 0.26\std{.03} (+0.03\std{.03})
  & 0.47\std{.03} (+0.06\std{.02})
  & 0.09\std{.01} (+0.03\std{.01}) \\
& AdaSteer  & 77.9\std{15.4}
  & \textbf{0.52\std{.06} (+0.02\std{.06})}
  & 0.23\std{.02} (+0.00\std{.00})
  & 0.41\std{.01} (+0.00\std{.00})
  & 0.06\std{.00} (+0.00\std{.00}) \\
& SADI      & 77.9\std{15.4}
  & 0.51\std{.07} (+0.01\std{.07})
  & 0.23\std{.01} (+0.00\std{.00})
  & 0.41\std{.01} (+0.00\std{.00})
  & 0.07\std{.00} (+0.00\std{.00}) \\
& ICD       & 77.9\std{15.4}
  & 0.51\std{.07} (+0.01\std{.07})
  & 0.27\std{.02} (+0.04\std{.00})
  & 0.44\std{.02} (+0.03\std{.01})
  & 0.07\std{.00} (+0.01\std{.00}) \\
\rowcolor{methodpurplelight} \cellcolor{white}
& \methodnameplain \textbf{(Ours)} & 77.9\std{15.4}
  & 0.50\std{.06} (+0.00\std{.06})
  & 0.36\std{.02} (+0.13\std{.03})
  & \textbf{0.51\std{.03} (+0.10\std{.04})}
  & \textbf{0.14\std{.02} (+0.07\std{.02})} \\
\midrule

\multirow{7}{*}{\textbf{Qwen3-4B}}
& Vanilla   & 0.0
  & 0.50\std{.00} & 0.33\std{.02} & 0.50\std{.02} & 0.11\std{.01} \\
& DoLa      & 45.4\std{3.9}
  & 0.72\std{.00} (+0.22\std{.00})
  & \textbf{0.76\std{.06} (+0.43\std{.06})}
  & 0.25\std{.07} (-0.25\std{.05})
  & 0.06\std{.03} (-0.05\std{.03}) \\
& ITI       & 45.4\std{3.9}
  & 0.60\std{.04} (+0.10\std{.04})
  & 0.35\std{.03} (+0.02\std{.02})
  & 0.52\std{.03} (+0.02\std{.01})
  & 0.11\std{.01} (+0.00\std{.01}) \\
& AdaSteer  & 45.4\std{3.9}
  & 0.75\std{.01} (+0.25\std{.01})
  & 0.33\std{.02} (+0.00\std{.00})
  & 0.50\std{.02} (+0.00\std{.00})
  & 0.11\std{.01} (+0.00\std{.00}) \\
& SADI      & 45.4\std{3.9}
  & 0.76\std{.01} (+0.26\std{.01})
  & 0.32\std{.02} (-0.01\std{.01})
  & 0.49\std{.02} (+0.00\std{.00})
  & 0.10\std{.01} (+0.00\std{.00}) \\
& ICD       & 45.4\std{3.9}
  & 0.73\std{.01} (+0.23\std{.01})
  & 0.36\std{.03} (+0.03\std{.02})
  & 0.53\std{.02} (+0.03\std{.01})
  & 0.11\std{.02} (+0.01\std{.01}) \\
\rowcolor{methodpurplelight} \cellcolor{white}
& \methodnameplain \textbf{(Ours)} & 45.4\std{3.9}
  & \textbf{0.78\std{.01} (+0.28\std{.01})}
  & 0.51\std{.03} (+0.18\std{.02})
  & \textbf{0.68\std{.01} (+0.19\std{.03})}
  & \textbf{0.23\std{.02} (+0.13\std{.01})} \\
\midrule

\multirow{7}{*}{\textbf{Mistral-7B}}
& Vanilla   & 0.0
  & 0.50\std{.00} & 0.34\std{.01} & 0.50\std{.01} & 0.13\std{.02} \\
& DoLa      & 58.3\std{6.0}
  & 0.69\std{.02} (+0.19\std{.02})
  & \textbf{0.71\std{.08} (+0.36\std{.08})}
  & 0.29\std{.09} (-0.21\std{.01})
  & 0.07\std{.04} (-0.06\std{.04}) \\
& ITI       & 58.3\std{6.0}
  & 0.44\std{.12} (-0.06\std{.12})
  & 0.31\std{.07} (-0.03\std{.06})
  & 0.47\std{.05} (-0.03\std{.06})
  & 0.10\std{.03} (-0.03\std{.02}) \\
& AdaSteer  & 58.3\std{6.0}
  & 0.69\std{.02} (+0.19\std{.02})
  & 0.34\std{.01} (+0.00\std{.00})
  & 0.50\std{.01} (+0.00\std{.00})
  & 0.13\std{.02} (+0.00\std{.00}) \\
& SADI      & 58.3\std{6.0}
  & 0.69\std{.01} (+0.19\std{.01})
  & 0.34\std{.02} (+0.00\std{.00})
  & 0.50\std{.01} (+0.00\std{.00})
  & 0.13\std{.02} (+0.00\std{.00}) \\
& ICD       & 58.3\std{6.0}
  & 0.64\std{.05} (+0.14\std{.05})
  & 0.35\std{.03} (+0.00\std{.02})
  & 0.51\std{.01} (+0.01\std{.01})
  & 0.15\std{.01} (+0.02\std{.01}) \\
\rowcolor{methodpurplelight} \cellcolor{white}
& \methodnameplain \textbf{(Ours)} & 58.3\std{6.0}
  & \textbf{0.72\std{.03} (+0.22\std{.03})}
  & 0.54\std{.04} (+0.20\std{.03})
  & \textbf{0.72\std{.05} (+0.22\std{.05})}
  & \textbf{0.29\std{.05} (+0.16\std{.05})} \\
\midrule

\multirow{7}{*}{\textbf{Llama-3.1-8B}}
& Vanilla   & 0.0
  & 0.50\std{.00} & 0.29\std{.01} & 0.44\std{.01} & 0.08\std{.01} \\
& DoLa      & 41.4\std{5.6}
  & 0.66\std{.02} (+0.16\std{.02})
  & \textbf{0.55\std{.12} (+0.27\std{.11})}
  & 0.29\std{.08} (-0.15\std{.09})
  & 0.03\std{.03} (-0.05\std{.03}) \\
& ITI       & 41.4\std{5.6}
  & 0.46\std{.01} (-0.04\std{.01})
  & 0.32\std{.02} (+0.03\std{.03})
  & 0.50\std{.02} (+0.05\std{.01})
  & 0.10\std{.01} (+0.02\std{.02}) \\
& AdaSteer  & 41.4\std{5.6}
  & 0.67\std{.01} (+0.17\std{.01})
  & 0.29\std{.01} (+0.00\std{.00})
  & 0.44\std{.01} (+0.00\std{.00})
  & 0.08\std{.01} (+0.00\std{.00}) \\
& SADI      & 41.4\std{5.6}
  & 0.66\std{.01} (+0.16\std{.01})
  & 0.29\std{.01} (+0.01\std{.00})
  & 0.44\std{.01} (+0.00\std{.00})
  & 0.08\std{.01} (+0.00\std{.00}) \\
& ICD       & 41.4\std{5.6}
  & \textbf{0.69\std{.00} (+0.19\std{.00})}
  & 0.37\std{.02} (+0.09\std{.01})
  & 0.53\std{.04} (+0.09\std{.03})
  & 0.15\std{.02} (+0.07\std{.02}) \\
\rowcolor{methodpurplelight} \cellcolor{white}
& \methodnameplain \textbf{(Ours)} & 41.4\std{5.6}
  & \textbf{0.69\std{.02} (+0.19\std{.02})}
  & 0.51\std{.09} (+0.22\std{.08})
  & \textbf{0.65\std{.11} (+0.21\std{.09})}
  & \textbf{0.25\std{.07} (+0.17\std{.07})} \\
\bottomrule
\end{tabular}
}
\vspace{-0.7cm}
\end{table}

\subsection{Additional benchmark and ablations}
\label{sec:ablation}
We further benchmark \pcnet against RAG-augmented generation, and we conduct an ablation on the training data size and the MLP projection dimensionality.
These experiments are conducted on Llama-3.2-1B and Mistral-7B across CoQA and TruthfulQA.

\begin{wrapfigure}{R}{0.5\textwidth}
    \setlength{\belowcaptionskip}{-5ex}
    \centering
    \includegraphics[width=\linewidth]{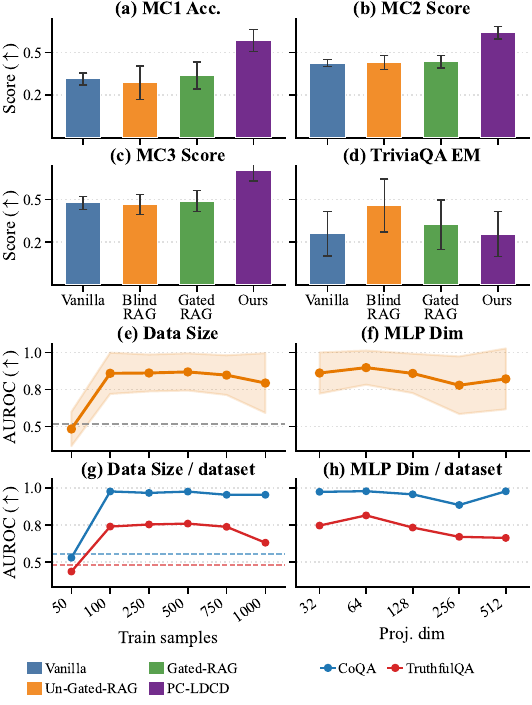}
    \caption{Results of the additional benchmark and ablations. \textbf{(a)}-\textbf{(d)} TruthfulQA MC1/MC2/MC3 and TriviaQA EM for vanilla, un-gated RAG, gated RAG, and \methodname (bars indicate mean over Llama-3.2-1B and Mistral-7B; error bars indicate the corresponding std). \textbf{(e)}, \textbf{(g)}: PCNet detection AUROC on CoQA and TruthfulQA as a function of training-set size, while \textbf{(f)}, \textbf{(h)} as a function of projection dimension (top-right, bottom-right); dashed lines mark the Token NLL baseline. \textbf{(e)}, \textbf{(f)} represent the average across datasets and LLMs, with the shadow being the std, whilst \textbf{(g)} and \textbf{(h)} represent the avg. across LLMs.}
    \label{fig:overall_ablation}
\end{wrapfigure}

\textbf{Benchmark against RAG-augmented generation}. To assess whether \pcnet's internal density signal is sufficient or external retrieval is necessary, \Cref{fig:overall_ablation} compares \methodname against Un-Gated RAG and Gated RAG across TruthfulQA-MC and TriviaQA (implementation details in \Cref{app:rag}).
\methodname substantially outperforms both baselines on all truthfulness metrics, MC1 ($0.570$ vs.\ $0.365$/$0.323$), MC2 ($0.616$ vs.\ $0.447$/$0.441$), MC3 ($0.669$ vs.\ $0.492$/$0.470$), while RAG retains an advantage on TriviaQA EM ($0.465$ vs.\ $0.290$), where retrieved passages directly contain the answer.
These results confirm that \pcnet is complementary to RAG: it achieves superior distributional truthfulness without retrieval, at the cost of exact-match recall on knowledge-intensive queries.

\setlength\intextsep{0pt}

\textbf{Ablation on training data size and MLP projection dimension.}
We ablate $n \in \{50, 100, 250, 500, 750, 1000\}$ training samples and projection dimensions $d \in \{32, 64, 128, 256, 512\}$, see \Cref{fig:overall_ablation}. 
On CoQA, \pcnet already reaches near-peak AUROC at $n{=}100$ (0.99 for Llama-3.2-1B, 0.97 for Mistral-7B); on the harder TruthfulQA setting, performance improves more gradually up to $n{=}750$ before slightly declining, suggesting mild overfitting at larger sizes. We adopt $n{=}500$ as the default, as the larger budget provides more stable training and guards against edge cases on harder benchmarks.
For the projection dimension, mean AUROC varies narrowly between $0.78$ and $0.86$ with no consistent trend, yet $d{=}128$ yields the highest latent alignment (mean cosine similarity $0.19$, fraction of positively-aligned dimensions $0.48$), indicating it best captures hallucination-relevant structure; we therefore fix $d{=}128$ throughout the experiments.
Full results are reported in \Cref{app:ablation} (\Cref{fig:training_data_abla,fig:mlp_ablation}).

% \textbf{Ablation on Training Data Size}. We ablate $n \in \{50, 100, 250, 500, 750, 1000\}$.
% On CoQA, \pcnet already reaches near-peak AUROC at $n{=}100$ (0.99 for Llama-3.2-1B, 0.97 for Mistral-7B); on the harder  TruthfulQA setting, performance improves more gradually up to $n{=}750$ before slightly declining, suggesting mild overfitting at larger sizes. 
% We adopt $n{=}500$ as the default: while near-optimal performance is already achieved at $n{=}100$, the larger budget provides a more stable training, guarding against edge cases on harder or lower-resource benchmarks.
% A full comparison is shown in \Cref{fig:overall_ablation} and under \Cref{app:ablation} in \Cref{fig:training_data_abla}.

% \textbf{Ablation on MLP Projection Dimension}. We ablate $d \in \{32, 64, 128, 256, 512\}$ CoQA and TruthfulQA: mean AUROC varies narrowly between $0.78$ and $0.86$ with no consistent trend, yet $d{=}128$ yields the highest latent alignment (mean cosine $0.19$, fraction of positively-aligned dimensions $0.48$), indicating it best captures hallucination-relevant structure.
% We therefore fix $d{=}128$ throughout all the experiments.
% A full comparison is shown in \Cref{fig:overall_ablation} and under \Cref{app:ablation} in \Cref{fig:mlp_ablation}.

%%%%%%%%%%%%%%%%%%%%%%%%%%%%%%%%%%%%%%%%%%%%%%%%%%%%%%%%%%%%

\section{Conclusions}
\label{sec:conclusions}

We introduced \pcnet, a tractable density estimator based on Probabilistic Circuits, and \methodname, a density-gated contrastive decoding strategy that together resolve the Detection-Correction Asymmetry in LLM hallucination mitigation.
By framing hallucinations as geometric anomalies on the factual manifold and computing exact NLL in a single forward pass, \pcnet achieves near-perfect detection across CoQA, SQuAD v2.0, and TriviaQA (AUROC up to $0.99$) without sampling, external verifiers, or weight modifications.

Crucially, using \pcnet as a dynamic gate prevents indiscriminate corrections from corrupting factual generations, reducing the mean corruption rate to $53.7\%$ and raising the preservation rate to $79.3\%$.
\methodname then routes interventions exclusively into the discrete token space, achieving the highest True+Info, MC2, and MC3 scores on TruthfulQA in three out of four models, while remaining competitive on standard QA benchmarks.
Ablation studies confirm robustness to projection dimensionality and strong data efficiency, with near-optimal detection already at $n{=}100$ training samples.
Qualitative correction examples are reported in \Cref{app:example}, while future directions and broader impacts are discussed respectively in \Cref{sec:future_works} and \Cref{app:broader_impact}.

\textbf{Limitations}. \pcnet requires a small calibration set of labeled factual and hallucinated hidden states, which may be costly in new domains. Moreover, our evaluation is limited to models up to 8B parameters, leaving open whether geometric separation holds in larger models. Finally, \methodname gains are stronger on generative benchmarks than on exact-match tasks, suggesting it is better suited to open-ended generation than precise answers.

%%%%%%%%%%%%%%%%%%%%%%%%%%%%%%%%%%%%%%%%%%%%%%%%%%%%%%%%%%%%

\clearpage

\bibliography{biblio}
\bibliographystyle{unsrtnat}

%%%%%%%%%%%%%%%%%%%%%%%%%%%%%%%%%%%%%%%%%%%%%%%%%%%%%%%%%%%%

\clearpage

\appendix
\crefalias{section}{appendix}

\section{Theoretical foundations}
\label{app:theoretical}
We provide the theoretical grounding for \pcnet and \methodname. We refer the reader to~\citep{darwiche2003diff} and~\citep{poon2011spn} for foundational treatments of probabilistic circuits, including formal treatments of smoothness, decomposability, and the linear-time inference guarantees we leverage; \citet{choi2020probabilistic} provides a modern unified framework, and~\citet{vergari2021compositional} systematizes tractable operations on circuits.

\subsection{Consistency of the contrastive manifold estimator}
\label{app:consistency}

\begin{proposition}[Consistency]
Let $h \sim p_{\text{factual}}$ denote a factual hidden state. Assume there exists $\phi^\star$ such that the density of $f_{\phi^\star}(h)$ lies in the PC family, and that the family is identifiable. Then under standard regularity conditions, the minimizers $(\hat\theta_N, \hat\phi_N)$ of $\mathcal{L}(\theta, \phi)$ over $N$ paired samples converge in probability to the population minimizer $(\theta^\star, \phi^\star)$ as $N \to \infty$ by standard M-estimator arguments.
\end{proposition}

\begin{remark}
The contrastive margin term acts as a bounded regularizer; under the stated assumptions, it does not alter the asymptotic consistency of the maximum-likelihood component, provided $\gamma > 0$ is held fixed.
\end{remark}

\subsection{No-regret on confident states}
\label{app:no_regret}

\begin{proposition}[No-regret on factual states]
\label{prop:gate-regret}
Define the gating strength $\beta_t = \sigma(\mathcal{S}_{\text{NLL}}(z_t) - \tau)$.
If $\mathcal{S}_{\text{NLL}}(z_t) \leq \tau - \log(1/\delta - 1)$, then $\beta_t \leq \delta$.
In this regime, the score assigned to every candidate $c_i$ in \eqref{eq:score} satisfies
\begin{equation}
    \bigl|\,\mathrm{Score}_{\text{LDCD}}(c_i) - 
\log P_{\text{LM}}(c_i \mid x_{<t})\bigr| 
\;\leq\; \delta \cdot \max_{c} 
|\mathcal{S}_{\text{NLL}}(f_\phi(h_{t+1}^{(c)}))|
\end{equation}
so \methodname is $\delta$-close to greedy decoding on states the \pcnet deems factual. In particular, for $\delta \to 0$, \methodname degenerates to the base LLM on high-confidence factual states, formally justifying the preservation rate guarantee.
\end{proposition}

% \begin{theorem}[Manifold-Preservation Bound]
% \label{thm:manifold}
% Suppose $-\log p_{\mathcal{C}}$ is $\beta$-smooth on the compact support of the calibration distribution and $f_\phi$ is $L_\phi$-Lipschitz. Let $z_t = f_\phi(h_t) \in \mathcal{M}_\epsilon$ and let $z_{t+1}^\star = f_\phi(h_{t+1}^{(x_t^\star)})$ be the latent code induced by the corrected token from \eqref{eq:score}. Denote by $\Delta h_t = h_{t+1}^{(x_t^\star)} - h_t$ the LLM's natural one-step transition. Then
% \begin{equation}
% \label{eq:manifold-bound}
% -\log p_{\mathcal{C}}(z_{t+1}^\star) \;\le\; -\log p_{\mathcal{C}}(z_t)\; +\; \langle \nabla_z (-\log p_{\mathcal{C}})(z_t),\, f_\phi(\Delta h_t)\rangle\; +\; \tfrac{\beta L_\phi^2}{2}\,\|\Delta h_t\|^2.
% \end{equation}
% The second-order term is bounded uniformly in $\Delta h_t$ by the LLM's per-token transition norm, and the first-order term is non-positive whenever the candidate $x_t^\star$ chosen by \eqref{eq:score} aligns with the score direction $\nabla_z \log p_{\mathcal{C}}(z_t)$. In particular, for sufficiently small transition norm, $z_{t+1}^\star \in \mathcal{M}_{2\epsilon}$.
% \end{theorem}

\subsection{Per-token computational overhead}
\label{app:complexity}

\begin{proposition}[Per-token overhead]
\label{prop:complexity}
At each generated token, \pcnet incurs three additive costs: (i) a single MLP encoder pass of cost $O(D_{LLM} \cdot D_{PC})$; (ii) a single PC forward pass of cost $O(|\mathcal{N}|)$, where $|\mathcal{N}|$ is the number of nodes in the circuit (see \Cref{def:pc} and~\citealp{darwiche2003diff}); (iii) for tokens flagged as anomalous only, $k$ partial lookahead passes through the final transformer block to materialize $\{h_{t+1}^{(c_i)}\}_{i=1}^k$. The amortized per-token overhead is therefore
\[
O\!\left(D_{LLM}\cdot D_{PC} \;+\; |\mathcal{N}| \;+\; \mathrm{IGR}\cdot k \cdot d_{\text{block}}\right),
\]
where $d_{\text{block}}$ is the cost of one final-block evaluation. For our default configuration ($|\mathcal{N}|\!\approx\!10^3$, $D_{LLM}\!=\!4096$, $D_{PC}\!=\!128$, $k\!=\!8$, $\mathrm{IGR} \in [0.41, 0.78]$ across models), the PC scoring cost is below $2\%$ of one full transformer-layer forward pass; the lookahead term dominates and scales as $\mathrm{IGR}\cdot k$ extra block evaluations per generated token.
\end{proposition}

The IGR values reported in our experiments (\Cref{tab:truthfulqa_results} and \Cref{tab:qa_benchmarks_results}) thus reflect the actual lookahead frequency, providing an empirical proxy for the wall-clock overhead of \methodname.

\section{Experimental setup}
\label{app:exp_set}

All experiments were conducted on a high-performance computing cluster using a single NVIDIA A100-SXM4 GPU with 40GB of VRAM, supported by 8 CPU cores and 35GB of system memory. 
Our implementation leverages \texttt{PyTorch} and \texttt{HuggingFace Transformers} within a \texttt{CUDA 12.5.0} environment.
The target LLM is frozen and loaded in 4-bit NormalFloat (NF4) quantization~\citep{dettmers2023qlora7} with FP16 compute dtype, ensuring memory efficiency without compromising the integrity of the latent representation manifolds extracted for \pcnet density estimation.

The \pcnet and its MLP bottleneck are trained jointly using the contrastive objective described in \Cref{sec:halu_dec}.
Following~\citep{du_haloscope_2024}, we reserve $25\%$ of available QA pairs per dataset for testing. 
For correction evaluation, performance is measured on a balanced subset of $300$ samples ($150$ factual, $150$ hallucinated), ensuring that corruption and preservation rates are computed on equal class support. All trials are run with three fixed RNG seeds ($42, 43, 44$) within a dedicated \texttt{Conda}  environment.
Full hyperparameters are reported in \Cref{tab:hyperparams}.

\begin{table}[!h]
\centering
\caption{Hyperparameters used for \pcnet training across all experiments.}
\label{tab:hyperparams}
\resizebox{0.6\columnwidth}{!}{%
\begin{tabular}{lc}
\toprule
\textbf{Hyperparameter} & \textbf{Value} \\
\midrule
Training samples         & 500 (250 factual / 250 hallucinated) \\
Epochs                   & 50 \\
Batch size               & 8 \\
Learning rate            & $10^{-3}$ \\
Weight decay             & $10^{-5}$ \\
Loss weight $\alpha$     & 0.8 \\
Margin $\gamma$          & 5.0 \\
Gradient clipping        & $\|\nabla\|_2 \leq 1.0$ \\
MLP projection dim $d$   & 128 \\
PC depth                 & 4 \\
PC branching factor      & 3 \\
LLM quantization         & 4-bit NF4 (FP16 compute) \\
RNG seeds                & 42, 43, 44 \\
\bottomrule
\end{tabular}}
\end{table}

\paragraph{Computational budget}
The full experimental pipeline comprises 216 model-training experiments across five experiment groups: baselines (48), \pcnet main runs (120), projection dimension ablations (16), RAG ablations (8), and data-size ablations (24).
All jobs were allocated 48 GPU-hours, each on a single NVIDIA A100-SXM4 (40GB), yielding a theoretical upper bound of $216 \times 48 = 10368$ GPU-hours.
Accounting for I/O overhead and average GPU utilization ($\sim$40\%), the effective compute is estimated at approximately $\mathbf{4147}$ GPU-hours.
The \pcnet runs reported in the main text represent the largest share ($5760$ allocated GPU-hours).

\subsection{Large Language Models, benchmark datasets, and evaluation metrics}

\paragraph{Large Language Models}
To evaluate the robustness of our framework across varying architectures and parameter scales, we conduct experiments on four LLMs spanning 1B to 8B parameters:
Llama-3.2-1B-Instruct~\citep{llama32herdofmodels}, Qwen3-4B~\citep{yang2025qwen3technicalreport},
Mistral-7B-v0.3~\citep{jiang2023mistral7b}, and Llama-3.1-8B-Instruct~\citep{llama32herdofmodels}.
This selection covers a broad range of model families, instruction-tuning regimes, and parameter scales, allowing us to assess whether \pcnet's geometric separation of factual and hallucinated hidden states transfers across architectures.

\paragraph{Benchmark Datasets}
We assess these models on four benchmarks targeting complementary factual challenges.
\textbf{CoQA}~\citep{reddy2019coqaconversationalquestionanswering} evaluates conversational reasoning over grounded passages, requiring the model to track dialogue context across turns.
\textbf{TriviaQA}~\citep{joshi-etal-2017-triviaqa} targets knowledge-intensive open-domain QA, where correct answers depend on parametric world knowledge rather than in-context evidence.
\textbf{SQuAD v2.0}~\citep{rajpurkar2018knowdontknowunanswerable} tests reading comprehension with unanswerable questions, probing the model's ability to abstain rather than hallucinate an answer. 
Finally, \textbf{TruthfulQA}~\citep{lin2022truthfulqameasuringmodelsmimic} specifically targets questions designed to elicit misconceptions encoded in pretraining weights, making it the most adversarial benchmark in our suite.

\paragraph{Evaluation Metrics}
We evaluate across three axes.
For \textbf{detection}, we report AUROC and F1, measuring the ability of \pcnet to separate hallucinated from factual hidden states at the instance level.
For \textbf{generation quality}, we use Exact Match (EM) on TriviaQA and SQuAD v2.0, and token-level F1 on CoQA.
On TruthfulQA, we report True+Info (T+I), MC1, MC2, and MC3: T+I jointly measures factual grounding and informativeness; MC1 measures single-answer accuracy against the most likely misconception; MC2 reports calibrated probability mass over all correct completions; and MC3 measures likelihood ranking against distractors. 
MC2 and MC3 are particularly diagnostic as they capture distributional truthfulness rather than single-answer confidence.
Finally, we report the \textbf{Instance Gating Rate} (IGR), i.e., the fraction of decoding steps where \pcnet triggers an intervention, alongside the \textbf{Corruption} rate (fraction of originally correct generations degraded by intervention) and the \textbf{Preservation} rate (fraction of correct generations successfully protected by gating), to jointly assess intervention selectivity and the safety of the gating mechanism.
In \Cref{tab:truthfulqa_results} and \Cref{tab:qa_benchmarks_results}, the IGR is identical across all correction methods within the same model; as all approaches share the same \pcnet gating signal, intervention decisions are made solely by \pcnet, independently of the downstream correction strategy applied.

\section{Detailed results}
\label{app:res_tables}

\Cref{tab:qa_benchmarks_results} reports the full correction performance on the standard QA benchmarks (CoQA, SQuAD v2.0, TriviaQA) in terms of token-level F1 and Exact Match (EM). Results are formatted as \textit{post-intervention score} ($\Delta$), where $\Delta$ denotes the change relative to the vanilla baseline, and are averaged across three RNG seeds. The IGR reflects the fraction of prompts where \pcnet actively triggered an intervention; a lower IGR indicates more selective gating, while a higher IGR reflects more pervasive anomaly detection on a given model--benchmark combination.

\begin{table}[!ht]
\centering
\caption{Performance on standard QA Benchmarks (CoQA, SQuAD v2.0,
TriviaQA). Results are formatted as \textit{score}\std{std} ($\Delta$),
where $\Delta$ denotes the change relative to Vanilla. IGR (\%) reflects the average Instance Gating Rate. Best post-intervention scores per model
are highlighted in \textbf{bold}. Averaged across three seeds.}
\label{tab:qa_benchmarks_results}
\resizebox{\textwidth}{!}{
\begin{tabular}{llcccc}
\toprule
\textbf{Model} & \textbf{Method} & \textbf{IGR (\%)} $\downarrow$
  & \textbf{CoQA F1} $\uparrow$
  & \textbf{SQuAD v2.0 EM} $\uparrow$
  & \textbf{TriviaQA EM} $\uparrow$ \\
\midrule

\multirow{7}{*}{\textbf{Llama-3.2-1B}}
& Vanilla  & 0.0  & 0.50 (+0.00) & 0.50 (+0.00) & 0.50 (+0.00) \\
& DoLa     & 77.9\std{15.4} & 0.68\std{.01} (+0.18) & 0.83\std{.00} (+0.33) & 0.58\std{.02} (+0.08) \\
& ITI      & 77.9\std{15.4} & 0.48\std{.00} (-0.02) & 0.49\std{.01} (-0.01) & 0.47\std{.01} (-0.03) \\
& AdaSteer & 77.9\std{15.4} & 0.70\std{.02} (+0.20) & \textbf{0.86\std{.01} (+0.36)} & \textbf{0.63\std{.00} (+0.13)} \\
& SADI     & 77.9\std{15.4} & 0.70\std{.02} (+0.20) & \textbf{0.86\std{.02} (+0.36)} & \textbf{0.63\std{.00} (+0.13)} \\
& ICD      & 77.9\std{15.4} & \textbf{0.71\std{.01} (+0.21)} & 0.83\std{.01} (+0.33) & 0.59\std{.01} (+0.09) \\
\rowcolor{methodpurplelight} \cellcolor{white}
& \methodnameplain \textbf{(Ours)} & 77.9\std{15.4}
  & 0.69\std{.02} (+0.19) & 0.84\std{.01} (+0.34) & 0.60\std{.01} (+0.10) \\
\midrule

\multirow{7}{*}{\textbf{Qwen3-4B}}
& Vanilla  & 0.0  & 0.50 (+0.00) & 0.50 (+0.00) & 0.50 (+0.00) \\
& DoLa     & 45.4\std{3.9} & 0.56\std{.02} (+0.06) & \textbf{0.90\std{.03} (+0.41)} & \textbf{0.70\std{.02} (+0.20)} \\
& ITI      & 45.4\std{3.9} & 0.53\std{.04} (+0.03) & 0.73\std{.12} (+0.23) & 0.51\std{.01} (+0.01) \\
& AdaSteer & 45.4\std{3.9} & \textbf{0.58\std{.02} (+0.08)} & \textbf{0.90\std{.02} (+0.41)} & 0.69\std{.02} (+0.19) \\
& SADI     & 45.4\std{3.9} & \textbf{0.58\std{.02} (+0.08)} & \textbf{0.90\std{.03} (+0.41)} & \textbf{0.70\std{.02} (+0.20)} \\
& ICD      & 45.4\std{3.9} & \textbf{0.58\std{.02} (+0.08)} & \textbf{0.90\std{.02} (+0.40)} & 0.67\std{.02} (+0.17) \\
\rowcolor{methodpurplelight} \cellcolor{white}
& \methodnameplain \textbf{(Ours)} & 45.4\std{3.9}
  & \textbf{0.58\std{.02} (+0.08)} & \textbf{0.90\std{.03} (+0.40)} & 0.67\std{.01} (+0.17) \\
\midrule

\multirow{7}{*}{\textbf{Mistral-7B}}
& Vanilla  & 0.0  & 0.50 (+0.00) & 0.50 (+0.00) & 0.50 (+0.00) \\
& DoLa     & 58.3\std{6.0} & 0.71\std{.01} (+0.21) & 0.84\std{.03} (+0.34) & 0.82\std{.01} (+0.32) \\
& ITI      & 58.3\std{6.0} & 0.53\std{.02} (+0.03) & 0.50\std{.01} (+0.00) & 0.50\std{.00} (+0.00) \\
& AdaSteer & 58.3\std{6.0} & 0.70\std{.01} (+0.20) & \textbf{0.86\std{.02} (+0.36)} & \textbf{0.84\std{.01} (+0.34)} \\
& SADI     & 58.3\std{6.0} & 0.71\std{.01} (+0.21) & \textbf{0.86\std{.02} (+0.36)} & \textbf{0.84\std{.01} (+0.34)} \\
& ICD      & 58.3\std{6.0} & 0.68\std{.01} (+0.18) & 0.67\std{.02} (+0.17) & 0.78\std{.02} (+0.28) \\
\rowcolor{methodpurplelight} \cellcolor{white}
& \methodnameplain \textbf{(Ours)} & 58.3\std{6.0}
  & \textbf{0.73\std{.01} (+0.23)} & 0.85\std{.02} (+0.35) & 0.81\std{.01} (+0.31) \\
\midrule

\multirow{7}{*}{\textbf{Llama-3.1-8B}}
& Vanilla  & 0.0  & 0.50 (+0.00) & 0.50 (+0.00) & 0.50 (+0.00) \\
& DoLa     & 41.4\std{5.6} & 0.72\std{.00} (+0.22) & 0.87\std{.02} (+0.37) & 0.79\std{.00} (+0.29) \\
& ITI      & 41.4\std{5.6} & 0.50\std{.01} (+0.00) & 0.50\std{.00} (+0.00) & 0.50\std{.00} (+0.00) \\
& AdaSteer & 41.4\std{5.6} & 0.70\std{.00} (+0.20) & 0.91\std{.02} (+0.41) & \textbf{0.82\std{.00} (+0.32)} \\
& SADI     & 41.4\std{5.6} & 0.71\std{.01} (+0.21) & \textbf{0.92\std{.02} (+0.42)} & \textbf{0.82\std{.00} (+0.32)} \\
& ICD      & 41.4\std{5.6} & \textbf{0.76\std{.01} (+0.26)} & 0.83\std{.02} (+0.33) & 0.76\std{.01} (+0.26) \\
\rowcolor{methodpurplelight} \cellcolor{white}
& \methodnameplain \textbf{(Ours)} & 41.4\std{5.6}
  & 0.69\std{.01} (+0.19) & 0.90\std{.03} (+0.40) & 0.79\std{.01} (+0.29) \\
\bottomrule
\end{tabular}
}
\end{table}

\section{Additional benchmark and ablations}
\label{app:ablation}

\subsection{Benchmark against RAG-augmented generation}
\label{app:rag}
The RAG ablation evaluates four approaches: \textit{Vanilla} (base LLM, no intervention), \textit{\methodname} (PCNet-gated correction), \textit{Un-Gated RAG} (retrieval always applied), and \textit{Gated RAG} (retrieval triggered only when \pcnet flags an anomaly). 
Retrieval is implemented via BM25 (Okapi, $k_1{=}1.5$, $b{=}0.75$) \citep{robertson2009probabilistic} over a corpus of up to $10000$ TriviaQA training passages, with top-$k{=}3$ passages pre-appended to the prompt as in-context evidence. 
For TruthfulQA-MC, all MC1 and MC2 choices are scored via teacher-forced mean log-probability; under Gated RAG, the retrieval-augmented prompt replaces the vanilla prompt only for flagged samples. 
The anomaly threshold $\tau$ is calibrated per model on the NLL distribution of the dataset's stored answers, selecting the value that maximizes the F1 score on a held-out validation split. 
For MC scoring, where ground-truth labels are unavailable at inference time, the median NLL across all choices is used as the threshold. All arms share the same frozen LLM and \pcnet checkpoint; only the prompt construction and decoding
strategies differ across conditions.

\subsection{Ablation on training data size}
\begin{figure}[!h]
\centering
    \includegraphics[width=0.8\textwidth]{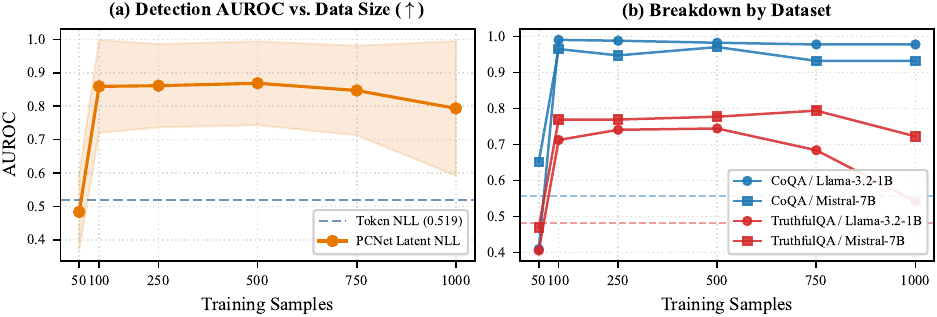}
    \caption{AUROC detection achieved by \pcnet on Llama 3.2-1B and Mistral-7B LLMs and on CoQa and TruthfulQA benchmark settings across different training dataset sizes. \textbf{(a)} The line represents the average across LLMs and datasets, and the shadow represents the standard deviation. \textbf{(b)} Each line refers to a single execution.}
    \label{fig:training_data_abla}
\end{figure}

The data-size ablation trains \pcnet with $n \in \{50, 100, 250, 500, 750, 1000\}$ balanced samples (50\% factual / 50\% hallucinated). For each size condition, the MLP projector and \pcnet are reinitialized from scratch while the base LLM remains frozen and is loaded only once per model--dataset pair. All other hyperparameters (depth$=4$, $d{=}128$, 50 epochs, lr$=10^{-3}$, margin$=5.0$, $\alpha{=}0.8$) are held fixed. Detection is evaluated on a held-out balanced test set of 200 samples using AUROC and F1.
The results are reported in \Cref{fig:training_data_abla}.

\subsection{Ablation on MLP projection dimension}
\begin{figure}[!h]
    \centering
    \includegraphics[width=0.8\textwidth]{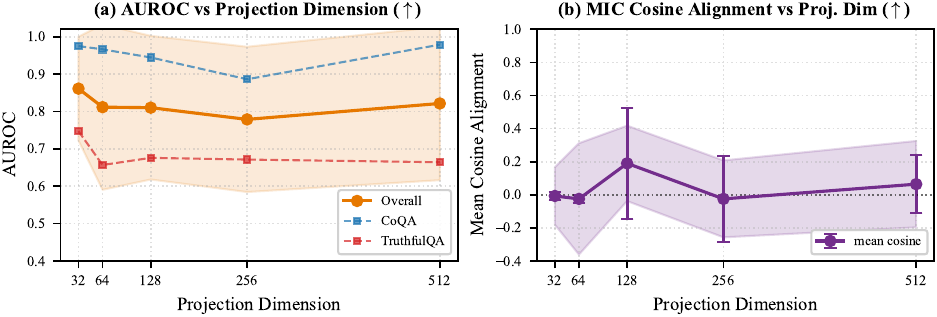}
    \caption{MLP projection dimensionality ablation across $d \in \{32, 64, 128, 256, 512\}$ on Llama-3.2-1B and Mistral-7B over CoQA and TruthfulQA.
    \textbf{(a)} Mean AUROC across all settings.
    \textbf{(b)} Latent alignment metrics (mean cosine similarity and fraction of positively-aligned dimensions), where $d{=}128$ achieves the highest alignment. Lines represent the mean across all settings, with shaded areas and error bars denoting the standard deviation.}
    \label{fig:mlp_ablation}
\end{figure}

The projection-dimension ablation sweeps $d \in \{32, 64, 128, 256, 512\}$ while keeping all other hyperparameters fixed as in \Cref{app:exp_set}.
For each value of $d$, the MLP projector and \pcnet are reinitialized independently. 
Detection performance is measured via AUROC; latent alignment quality is assessed via mean cosine similarity and the fraction of positively-aligned dimensions between the correction direction and the factual manifold, computed over held-out hidden states.
The results are reported in \Cref{fig:mlp_ablation}.

\setlength\intextsep{2ex}
\setlength{\belowcaptionskip}{3ex} 

\clearpage

\section{Qualitative examples of \methodname gated correction}
\label{app:example}

\noindent We present comprehensive qualitative examples of our gated correction mechanism in action. The examples are categorized by whether the model correctly detected and fixed a hallucination (Successfully Corrected), failed to fix it (Correction Failed), successfully preserved a correct answer (Not Corrected - Preserved), or incorrectly intervened (Corrupted).

\subsubsection*{Example 1 --- \textbf{Mistral-7B} on \textit{TriviaQA}}
\begin{itemize}
    \item \textbf{Context}: Who was Henry VIII's second wife?
    \item \textcolor{red}{\textbf{Answer}}: PARIS
    \item \textcolor{green!80!black}{\textbf{Correction}}: Anne Boleyn. \textit{(Successfully Corrected)}
\end{itemize}

\subsubsection*{Example 2 --- \textbf{Llama-3.1-8B} on \textit{TriviaQA}}
\begin{itemize}
    \item \textbf{Context}: Which other car company owns Seat?
    \item \textcolor{red}{\textbf{Answer}}: Borgia
    \item \textcolor{green!80!black}{\textbf{Correction}}: Volkswagen Group owns Seat. \textit{(Successfully Corrected)}
\end{itemize}

\subsubsection*{Example 4 --- \textbf{Qwen3-4B} on \textit{TriviaQA}}
\begin{itemize}
    \item \textbf{Context}: What is thalassophobia a fear of?
    \item \textcolor{red}{\textbf{Answer}}: Jonathan Swift
    \item \textcolor{green!80!black}{\textbf{Correction}}: Thalassophobia is a fear of the sea. It is a specific phobia characterized by an intense and irrational fear of oceans, water, or the sea. People with thalassophobia may experience anxiety, panic attacks, or avoidance behaviors when near the ocean or water. This fear can stem from various psychological, cultural, or personal experiences. \textit{(Successfully Corrected)}
\end{itemize}

\subsubsection*{Example 5 --- \textbf{Mistral-7B} on \textit{TriviaQA}}
\begin{itemize}
    \item \textbf{Context}: What is the nickname of Sheffield United
    \item \textcolor{red}{\textbf{Answer}}: Hugh Quarshie
    \item \textcolor{red}{\textbf{Correction}}: United \textit{(Correction Failed)}
\end{itemize}

\subsubsection*{Example 6 --- \textbf{Qwen3-4B} on \textit{TriviaQA}}
\begin{itemize}
    \item \textbf{Context}: Brown Willy is the highest point in which county?
    \item \textcolor{red}{\textbf{Answer}}: ROBERT SCHUMANN
    \item \textcolor{red}{\textbf{Correction}}: Brown Willy is the highest point in the county of Cornwall, England. It is a hill located in the south-western part of England, specifically in the area known as the Dartmoor region. The hill stands at an elevation of approximately 290 meters (951 feet) above sea level. It is part of the Dartmoor National Park, which is known for its \textit{(Correction Failed)}
\end{itemize}

\subsubsection*{Example 7  --- \textbf{Llama-3.2-1B} on \textit{TriviaQA}}
\begin{itemize}
    \item \textbf{Context}: Who led the quest for the golden fleece
    \item \textcolor{red}{\textbf{Answer}}: Rome
    \item \textcolor{red}{\textbf{Correction}}: None of them. The quest for the golden fleece was led by Philip II of Macedon. He was a king of Macedon in Greece. He led a series of expeditions to search for it. The story of the quest for the golden fleece is from the Iliad by Homer. It is a part of the epic poem that tells the story of the Greeks and their battles against the Tro \textit{(Correction Failed)}
\end{itemize}

\subsubsection*{Example 9 --- \textbf{Llama-3.2-1B} on \textit{TriviaQA}}
\begin{itemize}
    \item \textbf{Context}: What is thalassophobia a fear of?
    \item \textcolor{green!80!black}{\textbf{Answer}}: The Sea
    \item \textcolor{green!80!black}{\textbf{Correction}}: The Sea \textit{(Not Corrected - Preserved)}
\end{itemize}

\subsubsection*{Example 10  --- \textbf{Mistral-7B} on \textit{TriviaQA}}
\begin{itemize}
    \item \textbf{Context}: What is thalassophobia a fear of?
    \item \textcolor{green!80!black}{\textbf{Answer}}: The Sea
    \item \textcolor{green!80!black}{\textbf{Correction}}: The Sea \textit{(Not Corrected - Preserved)}
\end{itemize}

\subsubsection*{Example 11  --- \textbf{Llama-3.2-1B} on \textit{TriviaQA}}
\begin{itemize}
    \item \textbf{Context}: Who was Henry VIII's second wife?
    \item \textcolor{green!80!black}{\textbf{Answer}}: Anne Boleyn
    \item \textcolor{green!80!black}{\textbf{Correction}}: Anne Boleyn \textit{(Not Corrected - Preserved)}
\end{itemize}

\subsubsection*{Example 12  --- \textbf{Llama-3.1-8B} on \textit{TriviaQA}}
\begin{itemize}
    \item \textbf{Context}: What is the county town of Mayo?
    \item \textcolor{green!80!black}{\textbf{Answer}}: CASTLE BAR
    \item \textcolor{green!80!black}{\textbf{Correction}}: CASTLE BAR \textit{(Not Corrected - Preserved)}
\end{itemize}

\subsubsection*{Example 13  --- \textbf{Qwen3-4B} on \textit{TriviaQA}}
\begin{itemize}
    \item \textbf{Context}: Botswana was known by what name until 1966?
    \item \textcolor{green!80!black}{\textbf{Answer}}: Bechuanaland
    \item \textcolor{red}{\textbf{Correction}}: Before 1966, Botswana was known as Bechuanaland. The region was under British administration, and it was referred to as Bechuanaland Protectorate. In 1966, following a referendum and independence movement led by the Batswana people, the territory officially became the Republic of Botswana. The name change from Bechuanaland to Botswana \textit{(Corrupted)}
\end{itemize}

\subsubsection*{Example 14  --- \textbf{Mistral-7B} on \textit{TriviaQA}}
\begin{itemize}
    \item \textbf{Context}: What is the county town of Mayo?
    \item \textcolor{green!80!black}{\textbf{Answer}}: CASTLE BAR
    \item \textcolor{red}{\textbf{Correction}}: Castlebar is the county town of County Mayo in Ireland. It is located in the west of Ireland and is the largest town in County Mayo. Castlebar is known for its vibrant culture, historical sites, and natural beauty. It is also a hub for commerce, education, and entertainment in the region. The town was founded in the late 18th century and has a rich \textit{(Corrupted)}
\end{itemize}

\subsubsection*{Example 15  --- \textbf{Mistral-7B} on \textit{TriviaQA}}
\begin{itemize}
    \item \textbf{Context}: Sometimes called the 'Invicta flag', which English county's flag is a white horse on a red background?
    \item \textcolor{green!80!black}{\textbf{Answer}}: KENT
    \item \textcolor{red}{\textbf{Correction}}: Hampshire's flag is known as the 'Invicta flag' and features a horse on red background. The flag is based on the arms granted to Hampshire in 1660, which depict a horse rampant. The name 'Invicta' comes from the Latin word meaning 'unconquered', and refers to Hampshire's status as one of only two counties \textit{(Corrupted)}
\end{itemize}

\clearpage

\section{Future works}
\label{sec:future_works}
Several promising directions emerge from this work.
First, our framework currently extracts the hidden state of the \textit{last} token as the density input; however, recent evidence suggests that early token representations may encode factual commitments before the generation unfolds~\citep{marks2024geometrytruthemergentlinear}.
A rigorous investigation of whether the \textit{first} latent token already carries sufficient factual signal---and whether this enables earlier, cheaper anomaly detection---could substantially reduce inference overhead while improving robustness.

Second, and most consequentially, the advent of long chain-of-thought reasoning models (e.g., DeepSeek-R1, o1-style models) introduces a structured intermediate space that our framework has not yet exploited.
We conjecture that individual \textit{reasoning steps} in the chain-of-thought trace can be independently scored by \pcnet, enabling the identification and targeted correction of the precise step where a hallucinated factual commitment first emerges---rather than correcting the final answer after the error has propagated.
This would transform \methodname from a token-level intervention into a \textit{step-level reasoning corrector}, with the potential to dramatically improve factual consistency in mathematical, scientific, and multi-hop reasoning tasks where error accumulation across steps is the primary failure mode.

Third, our ablation against RAG reveals a complementary performance profile: \methodname achieves superior distributional truthfulness while RAG excels at knowledge-lookup tasks with explicit retrieved evidence. A natural extension is therefore a hybrid architecture of the two approaches.

Finally, scaling our evaluation to models beyond 8B parameters and extending \pcnet to multilingual and multimodal residual streams remain important open directions for establishing the broader applicability of latent density estimation as a foundation for trustworthy LLM inference.

\section{Broader impact}
\label{app:broader_impact}
By improving the factual reliability of LLMs, our framework contributes positively to the deployment of trustworthy AI systems in high-stakes domains, potentially reducing the spread of generated misinformation. However, as with any representation engineering technique, there is a risk of dual-use: the gating and manifold-learning mechanisms could theoretically be manipulated to enforce a biased or ideologically skewed ``factual'' manifold. Future work must ensure that the calibration datasets used for density estimation are diverse, transparent, and rigorously evaluated for representation biases.

\clearpage

\end{document}